\documentclass[runningheads]{llncs}

\usepackage{eccv}

\usepackage{eccvabbrv}

\usepackage{graphicx}
\usepackage{booktabs}

\usepackage[accsupp]{axessibility}  %

\usepackage{hyperref}

\usepackage{orcidlink}

\usepackage{url}

\usepackage[utf8]{inputenc} %
\usepackage[T1]{fontenc}    %
\usepackage{hyperref}       %
\usepackage{url}            %
\usepackage{booktabs}       %
\usepackage{amsfonts}       %
\usepackage{nicefrac}       %
\usepackage{microtype}      %
\usepackage{xcolor}         %

\usepackage{comment}
\usepackage{multirow}
\usepackage{graphicx}
\usepackage{enumitem} %
\usepackage{amsmath}

\usepackage{caption} %
\usepackage{float} %
\usepackage{bm}

\usepackage{subcaption}
\usepackage{tabularx}
\usepackage{makecell}
\usepackage{pifont}

\newcommand{\boldstartspace}[1]{\vspace{0.2em}\noindent\textbf{#1}}

\begin{document}

\title{ReSplat: Learning Recurrent Gaussian Splatting}

\author{Haofei Xu\inst{1,2} \and
Daniel Barath\inst{1} \and
Andreas Geiger\inst{2,3} \and Marc Pollefeys\inst{1,4}}

\authorrunning{H.~Xu et al.}

\institute{\textsuperscript{1}ETH Zurich \quad \textsuperscript{2}University of Tübingen, Tübingen AI Center \quad \textsuperscript{3}KE:SAI \quad \textsuperscript{4}Microsoft}

\maketitle

\begin{figure}[h!]
    \centering
    \vspace{-6mm}
    \includegraphics[width=\linewidth]{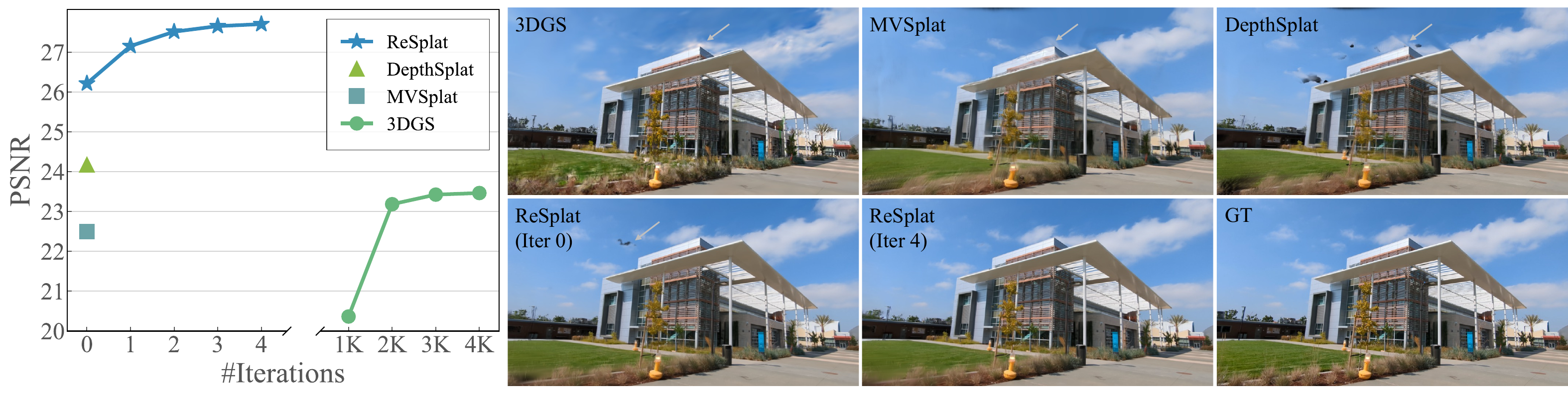}
    \caption{\textbf{Learning recurrent Gaussian splatting in a feed-forward manner}. 
    ReSplat iteratively refines 3D Gaussian Splatting (3DGS)~\cite{Kerbl2023TOG} for sparse view synthesis (2--32 views), a challenging regime where optimization-based 3DGS typically struggles. For initialization (iteration 0), we introduce a compact reconstruction model that predicts Gaussians in a $16\times$ subsampled space. This yields $16\times$ fewer Gaussians and $4\times$ faster rendering compared to per-pixel baselines MVSplat~\cite{chen2024mvsplat} and DepthSplat~\cite{xu2025depthsplat}. 
    The reduced Gaussian count enables highly efficient subsequent refinement.
    Compared to the optimization-based 3DGS~\cite{Kerbl2023TOG}, ReSplat is $100\times$ faster thanks to its feed-forward design, while maintaining the benefits of iterative updates.
    Here we show the results for 8 input views ($512 \times 960$ resolution) on the DL3DV~\cite{ling2023dl3dv} dataset (metrics in \cref{tab:highres_8view_dl3dv}).
    }
    \label{fig:teaser}
    \vspace{-8mm}
\end{figure}

\begin{abstract}
  While existing feed-forward Gaussian splatting models offer computational efficiency and can generalize to sparse view settings, their performance is fundamentally constrained by relying on a single forward pass for inference. We propose ReSplat, a feed-forward recurrent Gaussian splatting model that iteratively refines 3D Gaussians without explicitly computing gradients. Our key insight is that the Gaussian splatting rendering error serves as a rich feedback signal, guiding the recurrent network to learn effective Gaussian updates. This feedback signal naturally adapts to unseen data distributions at test time, enabling robust generalization across datasets, view counts, and image resolutions.
To initialize the recurrent process, we introduce a compact reconstruction model that operates in a $16 \times$ subsampled space, producing $16 \times$ fewer Gaussians than previous per-pixel Gaussian models. 
This substantially reduces computational overhead and allows for efficient Gaussian updates.
Extensive experiments across varying number of input views (2, 8, 16, 32), resolutions ($256 \times 256$ to $540 \times 960$), and datasets (DL3DV, RealEstate10K, and ACID) demonstrate that our method achieves state-of-the-art performance while significantly reducing the number of Gaussians and improving the rendering speed.
Our project page is at \href{https://haofeixu.github.io/resplat/}{haofeixu.github.io/resplat}.
\end{abstract}

\section{Introduction}
\label{sec:intro}

Feed-forward Gaussian splatting~\cite{charatan2023pixelsplat,Szymanowicz2024CVPR} aims to directly predict 3D Gaussian parameters from input images, eliminating the need for expensive per-scene optimization~\cite{Kerbl2023TOG} and enabling high-quality sparse-view reconstruction and view synthesis~\cite{chen2024mvsplat,liu2024mvsgaussian,zhang2024gaussian,wang2024freesplat}. Very recently, significant progress has been made in this line of research: feed-forward models~\cite{zhang2024gs,xu2025depthsplat,chen2025llrm,ye2024no,chen2024mvsplat360} can now produce promising reconstruction and view synthesis results from sparse input views.

Despite these advances, performance remains largely concentrated on standard in-domain benchmarks~\cite{zhou2018stereo,ling2023dl3dv}, and existing feed-forward models often struggle to generalize to unseen scenarios. Most current methods~\cite{charatan2023pixelsplat,chen2024mvsplat,zhang2024gs,xu2025depthsplat,chen2025llrm} learn a single-step mapping from images to 3D Gaussians, an approach inherently limited by network capacity in complex scenes.
In contrast, per-scene optimization~\cite{Kerbl2023TOG} achieves high-quality results via iterative updates but is computationally expensive, requiring thousands of gradient-based updates. This motivates our approach: using learned recurrent steps to progressively improve reconstruction, balancing feed-forward efficiency with adaptability of iterative optimization.

We identify that the rendering error provides a valuable feedback signal, informing the model about the quality of its prediction. 
This allows the network to adapt to the test data, reducing the dependence on the training distribution and leading to robust generalization. Furthermore, by iterating this process, the model incrementally refines its prediction.
This recurrent mechanism reduces learning difficulty by decomposing the task and increases model expressiveness with each update, effectively simulating a deeper, unrolled network~\cite{bai2019deep,geiping2025scaling}.

Driven by this observation, we begin with a single-step feed-forward Gaussian reconstruction model to initialize the recurrent process and then perform recurrent updates to improve the initial Gaussians. 
Since the recurrent updates occur in 3D space, where a large number of Gaussians would impose a significant computational burden, we design our initial model to predict Gaussians in a $16 \times$ subsampled space. 
This contrasts with most existing feed-forward models~\cite{charatan2023pixelsplat,Szymanowicz2024CVPR,szymanowicz2024flash3d,zhang2024gs} that predict one Gaussian per pixel, which scales poorly with increasing number of views and image resolutions. Our method achieves a $16\times$ reduction in the number of Gaussians while maintaining performance.

Based on this compact initial reconstruction, we train a weight-sharing recurrent network that iteratively improves the initial prediction. Crucially, the network leverages the rendering error of the input views to determine how to update the Gaussians. 
Specifically, we render the input views (available at test time) using the current prediction, compute the rendering error, and propagate it to the 3D Gaussians. 
The recurrent network then predicts the parameter updates from this error and the current Gaussians, without explicit gradient computation.

We validate our method through extensive experiments across diverse scenarios. On the challenging DL3DV~\cite{ling2023dl3dv} dataset, using 8 input views at $512 \times 960$ resolution, our learned recurrent model improves PSNR by 3.5dB, while using only $1/16$ of the Gaussians and achieving $4\times$ faster rendering speed. We also demonstrate that our recurrent model leads to robust generalization to unseen datasets, view counts, and image resolutions, where previous single-step feed-forward models usually struggle. With 16 input views at $540 \times 960$ resolution, we outperform Long-LRM~\cite{chen2025llrm} by 0.8dB PSNR while using $4\times$ fewer Gaussians. On the commonly used two-view RealEstate10K~\cite{zhou2018stereo} and ACID~\cite{liu2021infinite} benchmarks, ReSplat also achieves state-of-the-art results, demonstrating its strong performance.

\vspace{-1pt}
\section{Related Work}
\vspace{-1pt}

\boldstartspace{Feed-Forward Gaussian Splatting}. Significant progress has recently been made in feed-forward Gaussian models~\cite{zhang2024gs,chen2025llrm,xu2025depthsplat,jiang2025anysplat,liu2025worldmirror,ye2025yonosplat}.
However, two major limitations persist: First, most existing feed-forward models predict one or multiple Gaussians for each pixel~\cite{charatan2023pixelsplat,szymanowicz2024flash3d,chen2025llrm,chen2024mvsplat,zhang2024gs}, which produces millions of Gaussians when handling many input views and/or high-resolution images and thus limits scalability. 
Second, most existing methods are developed with single-step feed-forward inference. While conceptually simple, the achievable quality is bounded by network capacity for challenging and complex scenes. 
In this paper, we overcome these two limitations by first reconstructing Gaussians in a $16\times$ subsampled space, and then performing recurrent Gaussian updates based on the rendering error, which significantly improves efficiency and quality.
Unlike SplatFormer~\cite{chen2024splatformer}, which introduces a \textit{single-step non-recurrent} refinement network for \textit{optimized} 3DGS parameters, we propose a \textit{weight-sharing recurrent} network to iteratively improve the results from a \textit{feed-forward initialization}. In addition, SplatFormer is evaluated only on object-centric datasets and it is non-trivial to make it work for complex scenes (it is 2dB PSNR worse than our method). In contrast, our ReSplat targets scene-level benchmarks, and we demonstrate the effectiveness of the rendering error as an informative feedback signal, which we find crucial but is missing in SplatFormer.

\boldstartspace{Learning to Optimize}. Many tasks in machine learning and computer vision can be formulated as minimization problems with an optimization objective, where the solutions are found by iterative gradient descent~\cite{andrychowicz2016learning,lucas1981iterative,sun2010secrets}. Modern approaches~\cite{ma2020deep,teed2020raft,metz2022velo,harrison2022closer} try to simulate the optimization process by iteratively updating an initial prediction with a weight-sharing network, which usually achieves superior results compared to single-step regression methods, especially for out-of-distribution generalization. In vision, such a framework has been successfully applied to optical flow~\cite{teed2020raft}, stereo matching~\cite{lipson2021raft,wen2025foundationstereo}, scene flow~\cite{teed2021raft}, SLAM~\cite{teed2021droid}, Structure-from-Motion~\cite{li2024megasam}, and Multi-View Stereo~\cite{wang2021itermvs}. 
Unlike prior work that often relies on feature correlations~\cite{teed2020raft} for the recurrent process, we investigate this paradigm for the feed-forward Gaussian splatting task and identify the Gaussian rendering error as an informative feedback signal.

\boldstartspace{Learning to Optimize for View Synthesis}. In the context of view synthesis, DeepView~\cite{flynn2019deepview} predicts multi-plane images with learned gradient descent, where explicit gradient computation is necessary. In addition, G3R~\cite{chen2024g3r} learns to iteratively refine 3D Gaussians with the guidance of the explicitly computed gradients. However, our method is gradient free. Moreover, G3R requires well-covered 3D points for initialization and struggles with sparse points, while we directly predict initial Gaussians from posed images, without requiring any initial 3D points. Like G3R, QuickSplat~\cite{liu2025quicksplat} also relies on gradient computation but focuses on surface reconstruction. Another related work LIFe-GOM~\cite{wen2025life} iteratively updates the 3D reconstruction, but it focuses on human avatars with a hybrid Gaussian-mesh 3D representation. In contrast, our method aims to improve the quality and generalization of feed-forward Gaussian splatting models for general scenes.

\begin{figure}[!t]
    \centering
    \vspace{-5pt}
    \includegraphics[width=\linewidth]{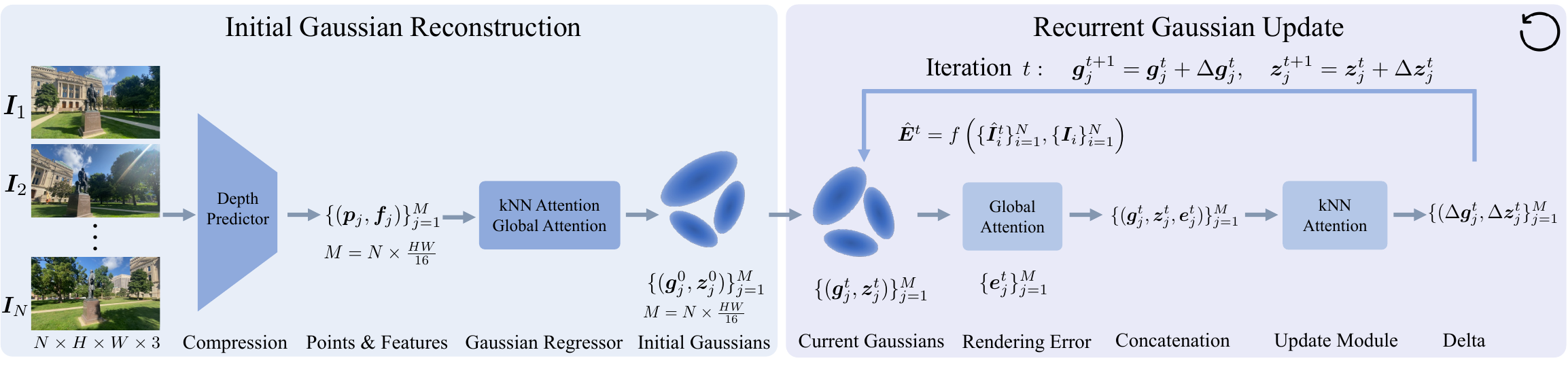}

    \caption{\textbf{ReSplat}. Given $N$ posed input images, we first predict per-view depth maps at $1/4$ resolution and then unproject and transform them into a point cloud with image features $\{({\bm p}_j, {{\bm f}}_j) \}_{j=1}^{M}$, where $M = N \times \frac{HW}{16}$ is the number of points. We then reconstruct an initial set of 3D Gaussians $\{ ({\bm g}^0_j, {\bm z}^0_j) \}_{j=1}^M$ in a $16\times$ subsampled 3D space using a $k$NN and global attention-based Gaussian regressor. Next, we learn to refine the initial Gaussians recurrently. At each recurrent step $t$, we use the current Gaussian prediction to render input views and then compute the rendering errors $\hat{\bm E}^t$ between rendered and ground-truth input views. Global attention is then applied on the rendering error to propagate these errors to the 3D Gaussians. A $k$NN attention-based update module then takes as input the concatenation of the current Gaussian parameters ${\bm g}^t_j$, the hidden state ${\bm z}^t_j$, and the rendering error ${\bm e}^t_j$, and predicts the incremental updates $\Delta {\bm g}^t_j $ and $\Delta {\bm z}^t_j$. We iterate this process for a total of $T$ steps. 
    }
    \label{fig:pipeline}
    \vspace{-4mm}
\end{figure}

\vspace{-4pt}
\section{Approach}
\vspace{-4pt}

Given $N$ input images $\{{\bm I}^{i}\}_{i=1}^N ({\bm I}^i \in \mathbb{R}^{H \times W \times 3}$) with their intrinsic $\{ \bm {K}^i \}_{i=1}^N$ ($\bm {K}^i \in \mathbb{R}^{3 \times 3}$) and extrinsic $\{ ({\bm R}_i, {\bm t}_i) \}_{i=1}^N$ (${\bm R}_i \in \text{SO}(3), {\bm t}_i \in \mathbb{R}^3 $) matrices, our goal is to predict a set of 3D Gaussian primitives~\cite{Kerbl2023TOG} $\mathcal{G} = \{ (\bm{\mu}_j, \alpha_j, \bm{\Sigma}_j, \bm{\mathrm{sh}}_j ) \}_{j=1}^M$ to model the scene, where $M$ is the total number of Gaussian primitives and $\bm{\mu}_j$, $\alpha_j$, $\bm{\Sigma}_j$, and $\bm{\mathrm{sh}}_j$ are the 3D Gaussian's position, opacity, covariance, and spherical harmonics, respectively. The reconstructed 3D Gaussians can be efficiently rasterized, enabling fast and high-quality novel view synthesis.

Unlike previous feed-forward models~\cite{charatan2023pixelsplat,zhang2024gs,chen2025llrm} that perform a single-step feed-forward prediction, we learn to estimate the Gaussian parameters recurrently. This not only reduces learning difficulty by decomposing the task into smaller, incremental steps but also enables higher reconstruction quality. In particular, we first predict an initial set of 3D Gaussians and then iteratively refine them in a gradient-free, feed-forward manner. Given that the Gaussian update occurs in 3D space, a large number of 3D Gaussians will introduce significant computational overhead during the update process. Thus, in our initial reconstruction stage, we predict a compact set of 3D Gaussians in a $16 \times$ subsampled space. More specifically, we perform $4\times$ spatial compression when predicting per-view depth maps, which leads to $16\times$ fewer Gaussians compared to previous per-pixel representations~\cite{charatan2023pixelsplat,Szymanowicz2024CVPR}. Consequently, the number of Gaussians $M$ in our model is $N \times \frac{HW}{16}$, which scales efficiently to many input views and high-resolution images. \cref{fig:pipeline} provides an overview of our pipeline.

\subsection{Initial Gaussian Reconstruction}
\label{sec:init_gaussian}

\boldstartspace{Subsampled 3D Space.} Our initial Gaussian reconstruction model is based on the DepthSplat~\cite{xu2025depthsplat} architecture. However, unlike DepthSplat, we predict Gaussians in a spatially $16\times$ subsampled 3D space ($N \times \frac{HW}{16}$), and thus we produce $16 \times$ fewer Gaussians than DepthSplat. To achieve $16 \times$ subsampling, we resize the full-resolution depth predictions from the depth model in DepthSplat to $1/4$ resolution ($N \times \frac{H}{4} \times \frac{W}{4}$), and then unproject and transform them into 3D via camera parameters to obtain a point cloud with $M = N \times \frac{HW}{16}$ points. Each 3D point ${\bm p}_j \in \mathbb{R}^3$ is also associated with a feature vector ${\bm f}_j \in \mathbb{R}^{C_0}$ ($C_0=256$ for our small model and $512$ for our base model) extracted from the input images: 
\begin{equation}
    \{{\bm I}_{i}, {\bm K}_i, {\bm R}_i, {\bm t}_i \}_{i=1}^N \rightarrow \{({\bm p}_j, {{\bm f}}_j) \}_{j=1}^{M}.
\end{equation}
Since we now have $16 \times$ fewer 3D points, na\"ively predicting Gaussian parameters from the point features ${{\bm f}}_j$ will lead to considerable performance loss. However, we find that using additional $k$NN attention~\cite{zhao2021point} and global attention~\cite{vaswani2017attention} layers to encode the 3D context~\cite{xu2023murf,chen2024hac} information can compensate for this loss.

\boldstartspace{Aggregating the 3D Context.} We use six alternating blocks of $k$NN attention and global attention to model both local and global 3D contexts, which enables communication between different 3D points and produces 3D context-aggregated features ${\bm f}^*_j \in \mathbb{R}^{C_0}$ with increased expressiveness:
\begin{equation}
    \{({\bm p}_j, {{\bm f}}_j) \}_{j=1}^{M} \rightarrow \{({\bm p}_j, {\bm f}^*_j) \}_{j=1}^{M}.
    \label{eq:init_pt}
\end{equation}

\boldstartspace{Decoding to Gaussians.} We use the point cloud $\{{\bm p}_j \}_{j=1}^{M}$ as the Gaussian centers, and other Gaussian parameters are decoded using a lightweight Gaussian head (two-layer MLP) from the 3D context-aggregated features $\{{\bm f}^*_j \}_{j=1}^{M}$. 
Accordingly, we obtain an initial set of 3D Gaussians with parameters $\{ (\bm{\mu}_j, \alpha_j, \bm{\Sigma}_j, \bm{\mathrm{sh}}_j) \}_{j=1}^{M}$ and feature vectors $\{ {\bm f}^*_j \}_{j=1}^{M}$. We use ${\bm g}^0_j \in \mathbb{R}^{C_1} (C_1=59)$ to denote the concatenation of all the Gaussian parameters $(\bm{\mu}_j, \alpha_j, \bm{\Sigma}_j, \bm{\mathrm{sh}}_j)$ for the $j$-th Gaussian at initialization, where $C_1$ is the total number of parameters for each Gaussian. We use ${\bm z}^0_j$ to denote the initial hidden state of the $j$-th Gaussian for the subsequent recurrent process, and initialize it with the feature ${\bm f}^*_j$: ${\bm z}^0_j = {\bm f}^*_j \in \mathbb{R}^{C_0}$. Thus, the initial Gaussians can be represented as
\begin{equation}
\label{eq:init_gaussian}
    \mathcal{G}^0 = \{ ({\bm g}^0_j, {\bm z}^0_j) \}_{j=1}^{M}.
\end{equation}

\subsection{Recurrent Gaussian Update}

Based on the initial Gaussian prediction in \cref{sec:init_gaussian} (\cref{eq:init_gaussian}), we train a recurrent network that iteratively refines the initial prediction. In particular, at iteration $t$ ($t=0, 1, \cdots, T-1$, where $T$ is the total number of iterations), the recurrent network predicts incremental updates to all Gaussian parameters $\Delta {\bm g}^t_j \in \mathbb{R}^{C_1}$ and their hidden state $\Delta {\bm z}^t_j \in \mathbb{R}^{C_0}$ as:
\begin{equation}
\label{eq:update}
    {\bm g}^{t+1}_j = {\bm g}^t_j + \Delta{\bm g}^t_j, \quad {\bm z}^{t+1}_j = {\bm z}^t_j + \Delta{\bm z}^t_j.
\end{equation}
To predict the incremental updates $\Delta {\bm g}^t_j$ and $\Delta {\bm z}^t_j$, we propose to learn the update in a gradient-free, feed-forward manner from the rendering error of input views.

\boldstartspace{Computing the Rendering Error}. Given that we have access to the input views at test time, we are able to create a feedback loop to guide the recurrent network to learn the incremental updates. Specifically, we first render the input views $\{\hat{\bm I}^t_i \}_{i=1}^N$ based on the current Gaussian parameters at iteration $t$ and then measure the difference between the rendered and ground-truth input views. We evaluate several different methods to compute the rendering error and observe that a combination of pixel-space and feature-space errors performs best.

In particular, we first use $\{\hat{\bm I}^t_i - {\bm I}_i \}_{i=1}^N$ to measure the rendering error in the pixel space, and then perform $4\times$ spatial downsampling with pixel unshuffle to align with the number of 3D Gaussians. For the feature-space rendering error, we extract the first three stage features (at $1/2$, $1/4$ and $1/8$ resolutions) of the ImageNet~\cite{deng2009imagenet} pre-trained ResNet-18~\cite{he2016deep} for the rendered input views and ground-truth input views, and bilinearly resize the three scale features to the same $1/4$ resolution, followed by concatenation. We denote the extracted features as $\{\hat{\bm F}^t_i \}_{i=1}^N$ and $\{{\bm F}_i \}_{i=1}^N$ ($\hat{\bm F}^t_i, {\bm F}_i \in \mathbb{R}^{\frac{H}{4} \times \frac{W}{4} \times C_2}$, where $C_2=256$) for the rendered and ground-truth input views, respectively. We then compute the difference between the features with subtraction $\{\hat{\bm F}^t_i - {\bm F}_i \}_{i=1}^N$. We combine pixel-space and feature-space rendering errors via element-wise addition. To match channel dimensions, the pixel-space error is first projected to the feature space using a linear layer followed by Layer Normalization~\cite{ba2016layer}.
This can be expressed as
\begin{equation}
    \hat{\bm E}^t = f\left( \{\hat{\bm{I}}_i^t\}_{i=1}^N, \{\bm{I}_i\}_{i=1}^N \right) = \{\hat{\bm F}^t_i - {\bm F}_i \}_{i=1}^N + \text{proj}(\{\hat{\bm I}^t_i - {\bm I}_i \}_{i=1}^N),
\end{equation}
where ``proj'' is the operation mentioned before to match dimensions. We denote all rendering errors as $\hat{\bm E}^t = \{\hat{\bm e}^t_j \}_{j=1}^{N \times \frac{H}{4} \times \frac{W}{4}}$, where $\hat{\bm e}^t_j \in \mathbb{R}^{C_2}$ is the $j$-th feature difference of dimension $C_2$ at iteration $t$.

\boldstartspace{Propagating the Rendering Error to Gaussians}. To propagate the rendering error to 3D Gaussians such that they can guide the network to update the Gaussians, a straightforward approach is to concatenate the rendering error $\hat{\bm e}^t_j$ with the Gaussians (${\bm g}^t_j, {\bm z}^t_j$) in a spatially aligned manner, since they have the same number of points ($N \times \frac{HW}{16}$). However, with this approach, the $j$-th Gaussian can only receive local information around the $j$-th rendered pixel, even though it can also contribute to other rendered pixels during the rendering process. To propagate the rendering error more effectively, we propose to apply global attention across all the $N \times \frac{H}{4} \times \frac{W}{4}$ rendering errors $\hat{\bm E}^t$, which enables each Gaussian to receive information from all rendering errors. This process can be formulated as follows:
\begin{equation}
    {\bm E}^t = \mathrm{global\_attention} (\hat{\bm E}^t) = \{{\bm e}^t_j \}_{j=1}^{N \times \frac{HW}{16}},
\end{equation}
where ${\bm e}^t_j$ is the $j$-th rendering error, which has aggregated the original point-wise rendering error $\hat{\bm e}^t_j$ globally. We then concatenate the Gaussians with the globally aggregated rendering errors as $\{ ({\bm g}^t_j, {\bm z}^t_j, {\bm e}^t_j) \}_{j=1}^{M}$, which are then used to predict the incremental update (illustrated in \cref{fig:pipeline}).

\boldstartspace{Recurrent Gaussian Update}. Letting the Gaussians at iteration $t$ be $\mathcal{G}^t = \{ ({\bm g}^t_j, {\bm z}^t_j) \}_{j=1}^{M}$, our update module predicts the incremental updates of Gaussian parameters and hidden state as:
\begin{equation}
\label{eq:recurrent}
    \{ ({\bm g}^t_j, {\bm z}^t_j, {\bm e}^t_j) \}_{j=1}^{M} \rightarrow \{ (\Delta{\bm g}^t_j, \Delta{\bm z}^t_j)\}_{j=1}^{M}.
\end{equation}
These updates are then added to the current prediction (\cref{eq:update}). This process is iterated $T$ times. We observe that our model converges after 4 iterations. During training, we randomly sample the
number of iterations $T$ between 1 and 4, and our model supports a different number of iterations at
inference time, allowing a flexible speed-accuracy trade-off with a single model. Since the recurrent process occurs in 3D space, we choose to use four $k$NN attention~\cite{zhao2021point} blocks as the recurrent architecture to model the local structural details. The Gaussian updates ${\bm g}^t_j$ are decoded with a lightweight head (four-layer MLP).

\subsection{Training Loss}

Our model is trained in two stages. In the first stage, we train an initial Gaussian reconstruction model to provide a compact initialization to our subsequent updates. The training loss is a combination of a rendering loss $\ell_{\mathrm{render}}$ and a depth smoothness loss~\cite{godard2019digging} $\ell_{\mathrm{smooth}}$ on the predicted depth maps of the input views:
\begin{eqnarray}
    \label{loss:1st}
    L_{\mathrm{1st}} & = & \sum_{v=1}^V \ell_{\mathrm{render}} (\hat{\bm I}_v, {\bm I}_v)  + \alpha \cdot \sum_{i=1}^N \ell_{\mathrm{smooth}} ({\bm I}_i, \hat{\bm D}_i), \\
    \label{loss:render}
    \ell_{\mathrm{render}} (\hat{\bm I}, {\bm I}) & = & \ell_1(\hat{\bm I}, {\bm I}) + \lambda \cdot \ell_{\mathrm{perceptual}}(\hat{\bm I}, {\bm I}),\\
\label{loss:depth_smooth}
    \ell_{\mathrm{smooth}} ({\bm I}, \hat{\bm D}) & = & |\partial_x \hat{\bm D}| e^{-|\partial_x \bm{I}|} + |\partial_y \hat{\bm D}| e^{-|\partial_y \bm{I}|},
\end{eqnarray}
where $V$ is the number of target views to render in each training step, and $N$ is the number of input views. The perceptual loss $\ell_{\mathrm{perceptual}}$~\cite{johnson2016perceptual} measures the distance in VGG~\cite{simonyan2014very} feature space, which is also used in previous methods~\cite{zhang2024gs,jin2024lvsm}. The depth smoothness loss $\ell_{\mathrm{smooth}}$ does not require ground-truth depth and serves as a regularization term on the estimated depth maps of the input views to encourage the depth gradient to be similar to the image gradient~\cite{godard2017unsupervised,godard2019digging}. We use $\alpha = 0.01$ and $\lambda=0.5$ for all the experiments.

In the second stage, we freeze our initial reconstruction model and train only the recurrent model end-to-end. 
We use the rendering loss $\ell_{\mathrm{render}}$ of rendered and ground-truth target views to supervise the network. 
All Gaussian predictions during the recurrent process are supervised using the rendering loss, applying exponentially ($\gamma=0.9$) increasing weights:
\begin{equation}
    L_{\mathrm{2nd}} = \sum_{t=0}^{T-1} \gamma^{T-1-t} \sum_{v=1}^V \ell_{\mathrm{render}} (\hat{\bm I}^t_v, {\bm I}_v).
\end{equation}

\vspace{-6pt}
\section{Experiments}
\vspace{-6pt}

\boldstartspace{Implementation Details.} We implement our method in PyTorch~\cite{paszke2019pytorch}. We choose $k=16$ for $k$NN attention in the initialization model following Point Transformer~\cite{zhao2021point}, and we use $k=8$ for the recurrent model to focus more on local details. Our Gaussian splatting renderer is based on the Mip-Splatting~\cite{yu2024mipsplatting} implementation in gsplat~\cite{ye2025gsplat}. We optimize our model with the AdamW~\cite{loshchilov2017decoupled} optimizer. More training details are presented in the appendix.

\textbf{Efficient Global Attention Implementation.} Our model contains several global attention layers. Considering that performing global attention on $N \times \frac{H}{4} \times \frac{W}{4}$ features would be expensive for high-resolution images, we first perform $4\times$ spatial downsampling with pixel unshuffle (reshaping from the spatial dimension to the channel dimension) and then compute global attention on the $N \times \frac{H}{16} \times \frac{W}{16}$ features. Finally, we upsample the features back to $1/4$ resolution using pixel shuffle (reshaping from the channel dimension to the spatial dimension). This implementation enables our model to scale efficiently to high-resolution images.

\textbf{Efficient $k$NN Implementation.} For typical point counts (e.g., $<300$K), we default to a customized CUDA implementation~\cite{pointcept2023} of global $k$NN. However, because this GPU implementation has $O(N^2)$ complexity, it becomes a computational bottleneck at higher resolutions or view counts. To ensure scalability in these regimes, we introduce an efficient local $k$NN alternative that reduces complexity to $O(N \cdot C)$ by restricting the search to a small candidate set of size $C$. This set is constructed using the known camera parameters to aggregate spatial neighbors from the same view and cross-view neighbors projected from nearby cameras. Computing exact 3D distances strictly within this constant-sized set guarantees efficient top-$k$ selection even for massive point clouds. We provide more computational analysis in the appendix.

\textbf{Model Sizes.} Our default model (ReSplat-Base) uses a ViT-B~\cite{dosovitskiy2020image,oquab2023dinov2,yang2024depth} backbone as part of our depth prediction model, which has 223M parameters in total (209M for the initialization model and 14M for the recurrent model). For ablation experiments, we use a ViT-S backbone (ReSplat-Small) to save compute, which has 77M parameters in total (62M for the initialization model and 15M for the recurrent model). In \cref{tab:recurrent_vs_better_init}, we additionally train a large initialization model (ReSplat-Large, 559M) using a ViT-L backbone to evaluate scalability.

To facilitate reproducibility, we release our code and pre-trained models publicly at \url{https://github.com/cvg/resplat}.

\boldstartspace{Coordinate System.} Since our recurrent network operates within a global 3D space, the selection of a coordinate system is critical, as it directly determines the spatial distribution of the Gaussian's centers. For our datasets, camera poses are estimated from COLMAP~\cite{schonberger2016structure}. We evaluated aligning the global reference frame to various views within the sparse input set. Empirically, we observed that using the spatially central (\eg, the middle frame in a sequential trajectory) input view as the reference coordinate system yields the best performance (see \cref{tab:ablation_coord_system}). We posit that this centers the coordinate system, reducing the maximum transformation distance to the most distant input views and effectively balancing the spatial positions of the 3D Gaussians.

\boldstartspace{Evaluation Settings.} We mainly consider three evaluation settings. First, we evaluate view synthesis from 8 input views at $512 \times 960$ resolution on the DL3DV~\cite{ling2023dl3dv} dataset, where we retrain 3DGS~\cite{Kerbl2023TOG}, MVSplat~\cite{chen2024mvsplat}, and DepthSplat~\cite{xu2025depthsplat} with their public code for fair comparisons. Second, we consider view synthesis from 16 input views at $540 \times 960$ resolution on DL3DV following Long-LRM~\cite{chen2025llrm}. Third, we evaluate on the commonly used 2-view ($256 \times 256$) setting on RealEstate10K~\cite{zhou2018stereo} and ACID~\cite{liu2021infinite}, where we compare with 2-view methods like GS-LRM~\cite{zhang2024gs} and LVSM~\cite{jin2024lvsm}.

\vspace{-3pt}
\subsection{Main Results}
\vspace{-3pt}

\begin{table}[!t]

\begin{center}
\footnotesize

\vspace{-5pt}
\caption{\textbf{Evaluation of 8 input views ($512 \times 960$) on DL3DV}. The standard optimization-based approach 3DGS~\cite{Kerbl2023TOG} requires several thousand iterations to reach convergence, while our feed-forward ReSplat is $100\times$ faster and benefits from additional iterations. In contrast to per-pixel feed-forward models such as MVSplat~\cite{chen2024mvsplat} and DepthSplat~\cite{xu2025depthsplat}, which produce millions of Gaussians, our ReSplat compresses the Gaussian count by $16 \times$, resulting in a $4\times$ faster rendering speed.
    }
    \vspace{-6mm}
    \setlength{\tabcolsep}{2pt} %
    \resizebox{\linewidth}{!}{
    \begin{tabular}{lcccccccccccccccccccccccc}
    \toprule
    Method & Category & \#Iterations & PSNR $\uparrow$ & SSIM $\uparrow$ & LPIPS $\downarrow$ & \#Gaussians & {\begin{tabular}[x]{@{}c@{}}Recon.\\Time (s) \end{tabular}} & {\begin{tabular}[x]{@{}c@{}}Render \\Time (s) \end{tabular}} \\
    
    \midrule
    \multirow{4}{*}{3DGS~\cite{Kerbl2023TOG}} & \multirow{4}{*}{Optimization} & 1000 & 20.36 & 0.667 & 0.448 & 9K & 15 & 0.0001 \\
    & & 2000 & 23.18 & 0.763 & 0.269 & 137K & 31 & 0.0005 \\
    & & 3000 & 23.42 & 0.770 & 0.232 & 283K & 50 & 0.0008 \\
    & & 4000 & 23.46 & 0.770 & 0.224 & 359K & 70 & 0.0009 & \\
    \midrule
    MVSplat~\cite{chen2024mvsplat} & Feed-Forward & 0 & 22.49 & 0.764 & 0.261 & 3932K & 0.129 & 0.0030 \\
    DepthSplat~\cite{xu2025depthsplat} & Feed-Forward & 0 & 24.17 & 0.815 & 0.208 & 3932K & 0.190 & 0.0030 \\
    \midrule
     \multirow{5}{*}{ReSplat} & \multirow{5}{*}{Feed-Forward} & 0 & 26.21 & 0.842 & 0.185 & 246K & 0.311 & 0.0007 \\
     &  & 1 & 27.15 & 0.859 & 0.169 & 246K & 0.437 & 0.0007 \\
     &  & 2 & 27.51 & 0.865 & 0.163 & 246K & 0.563 & 0.0007 \\
     &  & 3 & 27.65 & 0.867 & 0.161 & 246K & 0.689 & 0.0007 \\
     &  & 4 & \textbf{27.70} & \textbf{0.868} & \textbf{0.160} & 246K & 0.816 & 0.0007 \\

    \bottomrule
    \end{tabular}
    }

    \label{tab:highres_8view_dl3dv}
    \end{center}
    \vspace{-4mm}
\end{table}

\begin{figure}[!t]
    \centering
    \includegraphics[width=\linewidth]{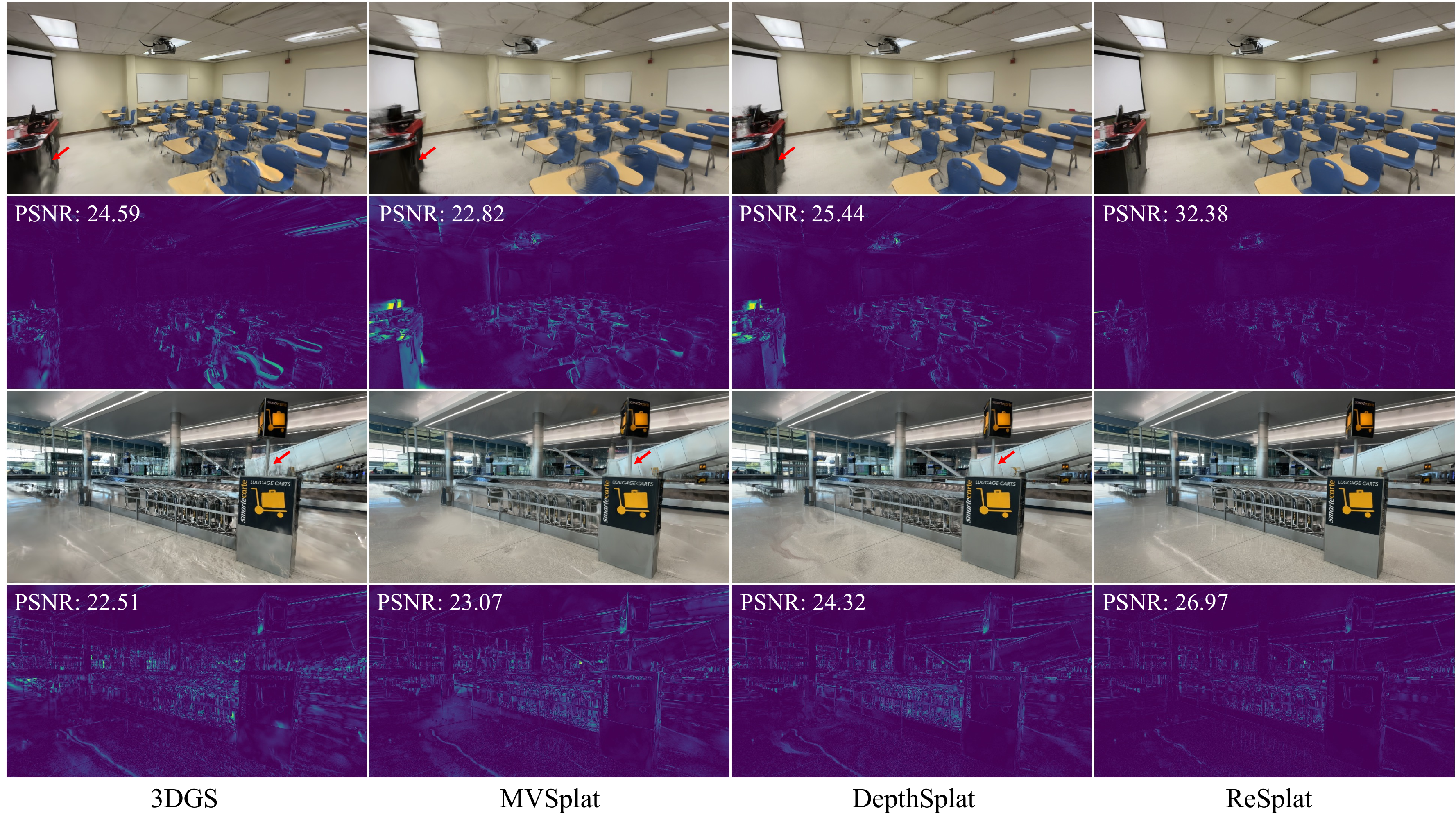}

    \vspace{-2mm}
    \caption{\textbf{View synthesis on DL3DV}. Our ReSplat outperforms both optimization-based and feed-forward methods, demonstrating significantly smaller rendering errors.
    }
    \label{fig:vis_compare}
    \vspace{-2mm}
\end{figure}

\boldstartspace{8 Views at $512 \times 960$ Resolution.} We report the results on the DL3DV~\cite{ling2023dl3dv} benchmark split (140 scenes) in \cref{tab:highres_8view_dl3dv}. Regarding 3DGS~\cite{Kerbl2023TOG}, we perform per-scene optimization on the 8 input views for all 140 scenes, while for feed-forward models, we perform zero-shot inference. We observe that 3DGS optimization typically converges with 4K optimization steps, and optimizing longer can lead to overfitting due to the sparse input views; thus, we report the best results (at 4K iterations). As shown in \cref{tab:highres_8view_dl3dv}, 3DGS optimization is computationally expensive due to the large number of iterations required, while our feed-forward ReSplat is $100 \times$ faster and is able to benefit from recurrent iterations. Previous per-pixel feed-forward models MVSplat~\cite{chen2024mvsplat} and DepthSplat~\cite{xu2025depthsplat} produce millions of Gaussians, while our ReSplat compresses the number of Gaussians by $16 \times$, resulting in a $4\times$ faster rendering speed. Overall, our ReSplat outperforms 3DGS by 4.2dB PSNR and DepthSplat by 3.5dB PSNR with superior efficiency on the number of Gaussians and the rendering speed. Visual comparisons provided in \cref{fig:vis_compare} and the appendix demonstrate the higher rendering quality of our method.

\begin{figure}[!t]
  \centering
  \begin{subfigure}[b]{0.49\textwidth}
    \centering
    \includegraphics[width=\textwidth]{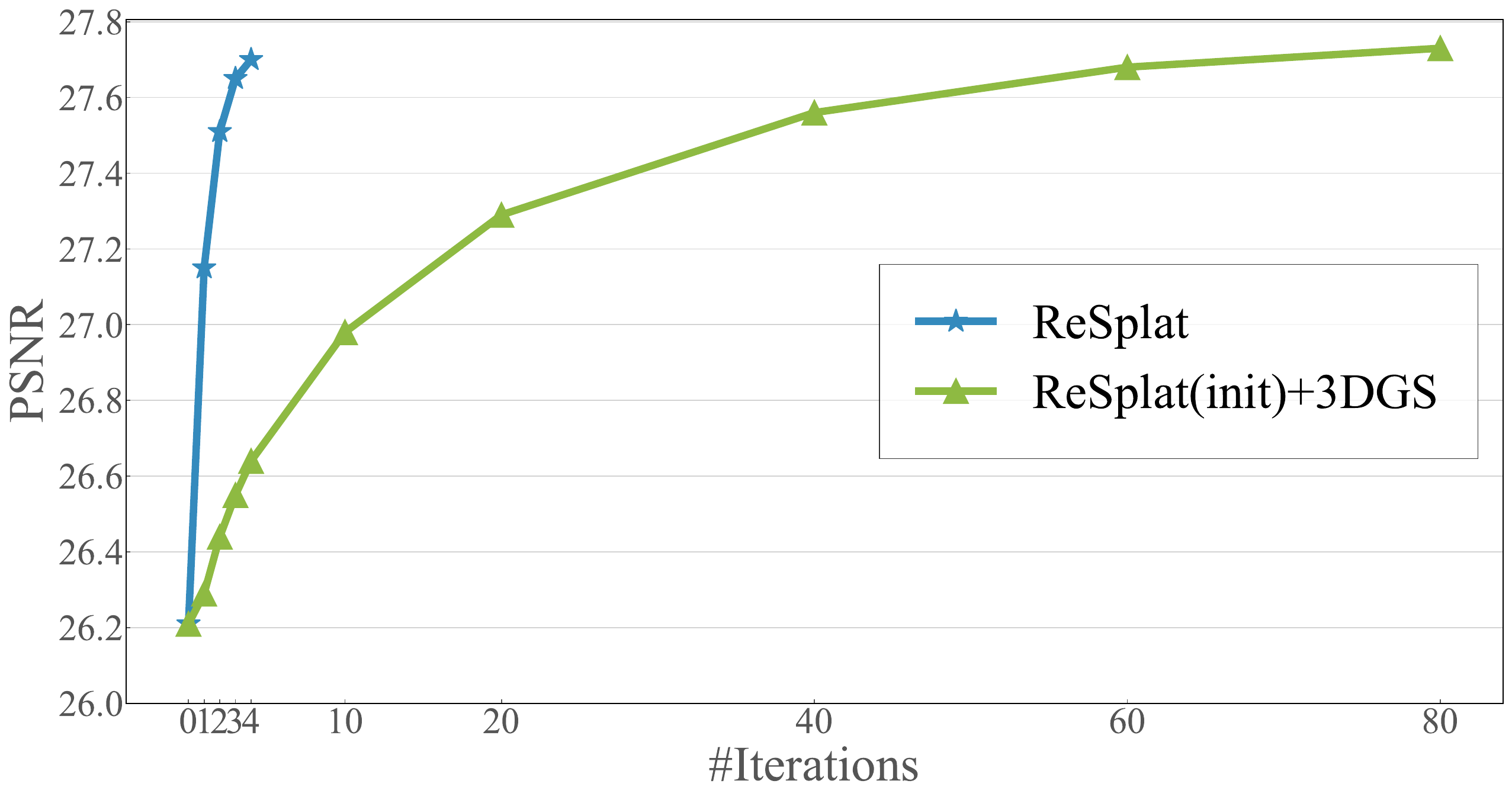}
    \caption{\textbf{PSNR \vs number of iterations.} 
    }
    \label{fig:3dgs_resplat_init_psnr}
  \end{subfigure}
  \begin{subfigure}[b]{0.49\textwidth}
    \centering
    \includegraphics[width=\textwidth]{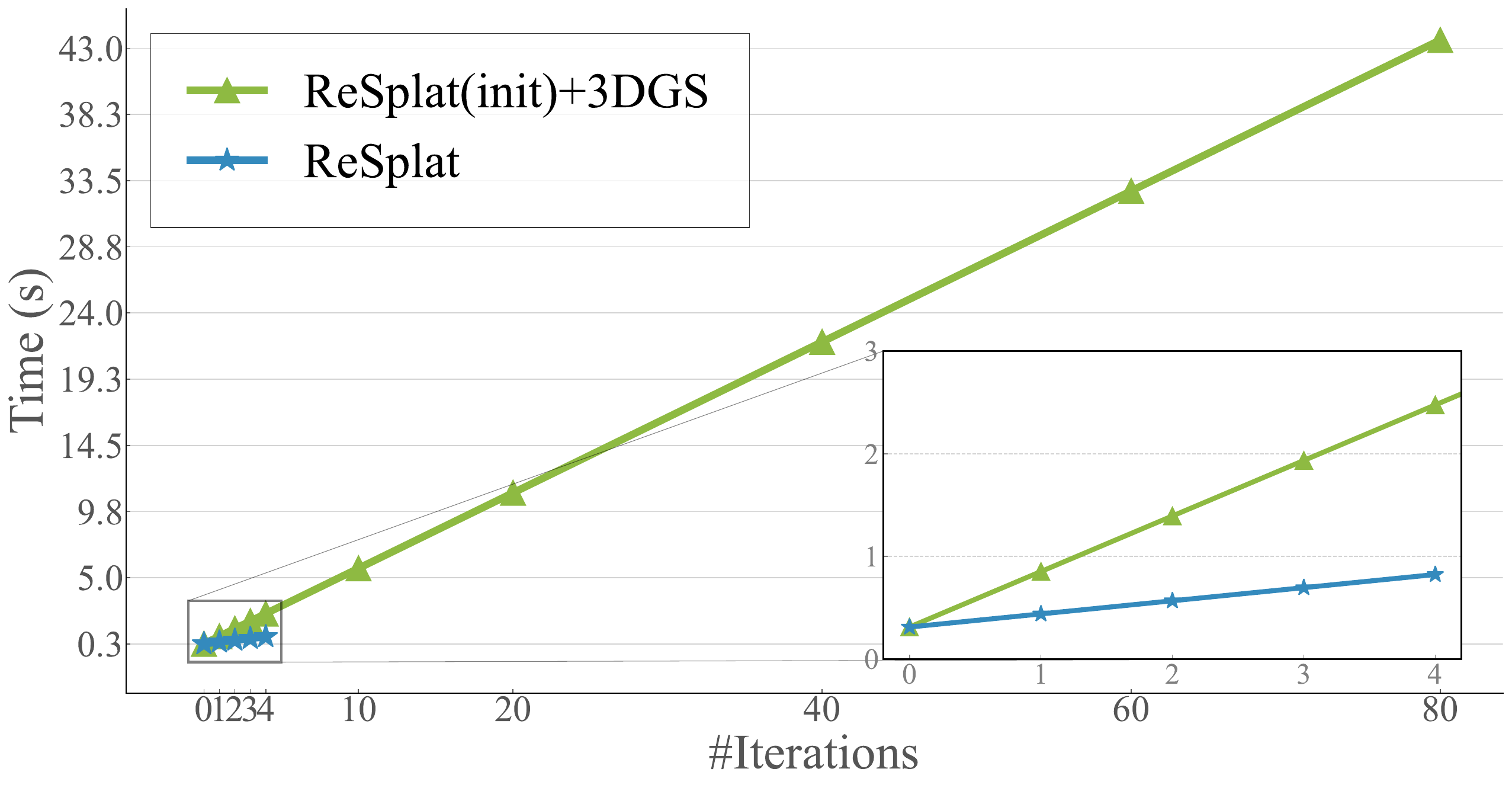}
    \caption{\textbf{Time \vs number of iterations.} 
    }
    \label{fig:3dgs_resplat_init_time}
  \end{subfigure}
  \vspace{-2mm}
  \caption{\textbf{Optimization-based \vs. feed-forward refinement.} Starting from the \textit{same ReSplat initialization}, we compare our feed-forward refinement against per-scene optimization using 3DGS~\cite{Kerbl2023TOG}. Our ReSplat improves the rendering quality significantly faster (4 \vs 80 iterations) and provides a $53\times$ speedup in reconstruction time. Furthermore, as highlighted in the zoomed-in region of (b), our per-iteration speed is faster than standard optimization since our approach eliminates the need for gradient computation.}
  \label{fig:3dgs_resplat_init}
  \vspace{-4mm}
\end{figure}

\boldstartspace{Optimization-Based \vs. Feed-Forward Refinement.} To further demonstrate the efficiency of our feed-forward refinement, we compare it against per-scene optimization using the \textit{same ReSplat initialization}. As illustrated in \cref{fig:3dgs_resplat_init}, our ReSplat is significantly faster than 3DGS optimization-based refinement thanks to our gradient-free, feed-forward architecture.

\begin{figure}[!t]
  \centering
  \begin{subfigure}[b]{0.32\textwidth}
    \centering
    \vspace{-4mm}
    \includegraphics[width=\textwidth, height=4cm, keepaspectratio]{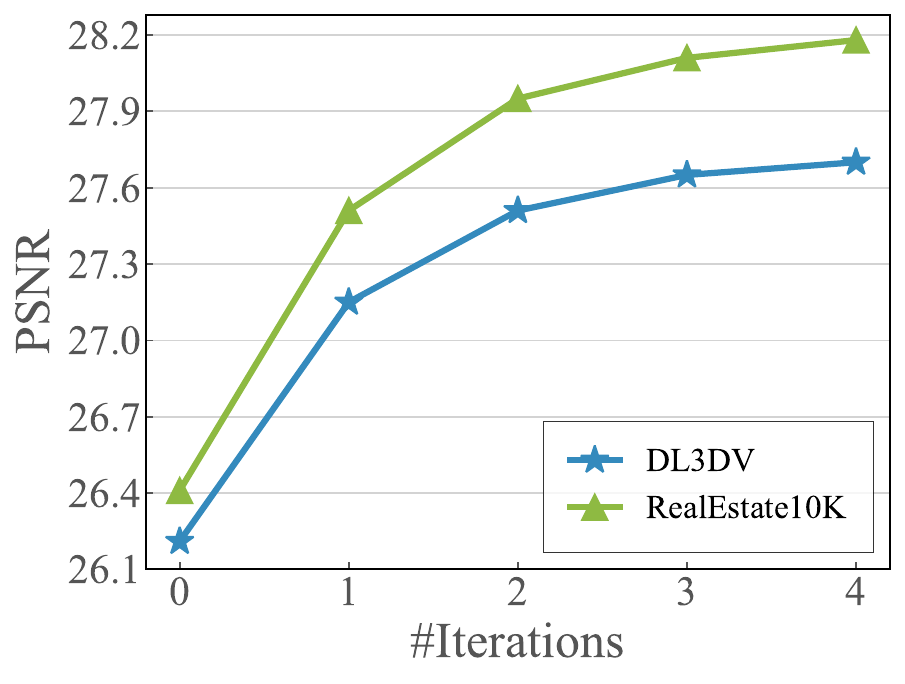}
    \vspace{-4mm}
    \caption{\textbf{Cross-dataset.} 
    }
    \label{fig:cross_data_gen}
  \end{subfigure}
  \begin{subfigure}[b]{0.32\textwidth}
    \centering
    \vspace{-4mm}
    \includegraphics[width=\textwidth, height=4cm, keepaspectratio]{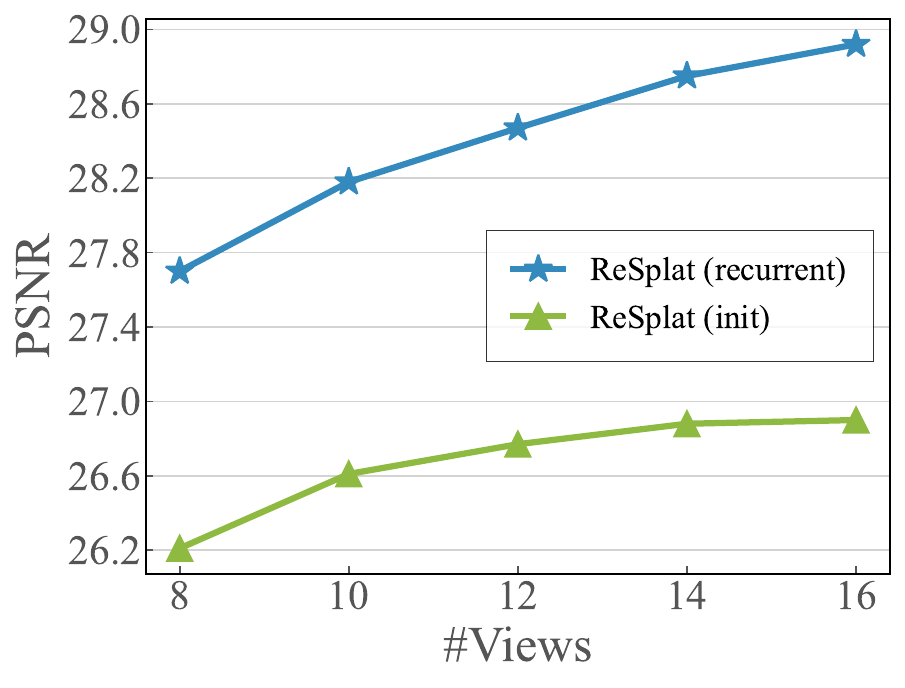}
    \vspace{-4mm}
    \caption{\textbf{Cross-view.} 
    }
    \label{fig:cross_view_gen}
  \end{subfigure}
  \begin{subfigure}[b]{0.33\textwidth}
    \centering
    \vspace{-4mm}
    \includegraphics[width=\textwidth, height=4cm, keepaspectratio]{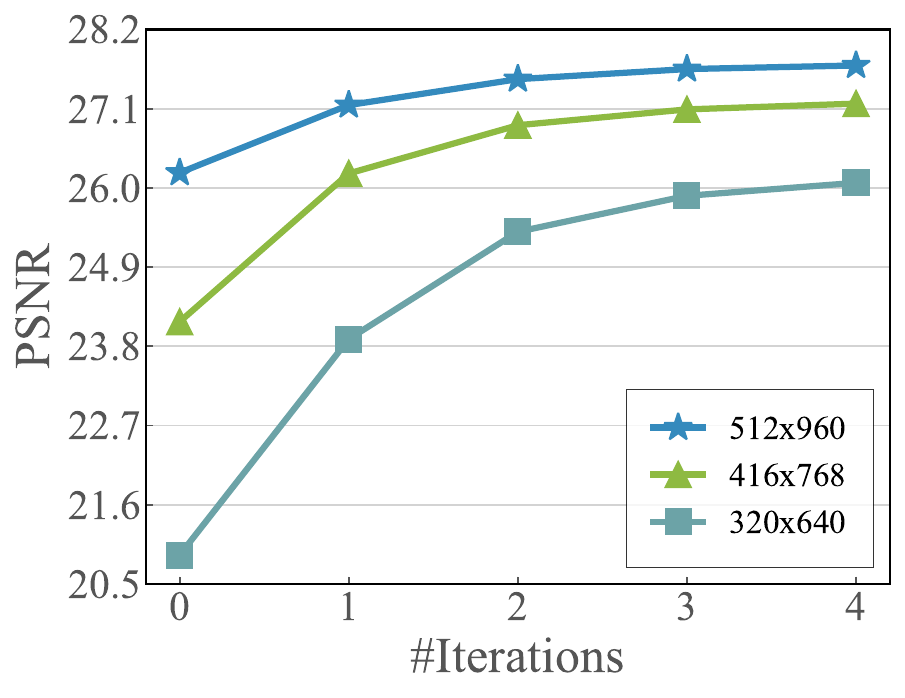}
    \vspace{-4mm}
    \caption{\textbf{Cross-resolution.} 
    }
    \label{fig:cross_res_gen}
  \end{subfigure}
  
  \caption{\textbf{Generalization to unseen datasets, view counts, and resolutions.} Our recurrent model demonstrates robust generalization capabilities, despite being trained solely on DL3DV at a fixed resolution ($512 \times 960$) with 8 input views. 
  }
  \label{fig:generalization}
\vspace{-4pt}
  
\end{figure}

\begin{table}[!t]
\begin{center}
\footnotesize
\caption{\textbf{Single-step \vs recurrent models.} Our recurrent ReSplat-Small (77M) outperforms all single-step baselines, including the significantly larger non-recurrent ReSplat-Large (559M), AnySplat (886M) and WorldMirror (1263M). This demonstrates that the benefits of recurrent error correction cannot be matched by simply increasing model parameters.}
\vspace{-2mm}
\setlength{\tabcolsep}{4pt}
\begin{tabular}{llccccc}
\toprule
 & Method & Params & PSNR $\uparrow$ & SSIM $\uparrow$ & LPIPS $\downarrow$ \\
\midrule
\multirow{5}{*}{Single-step} & AnySplat~\cite{jiang2025anysplat} & 886M & 20.79 & 0.643 & 0.241 \\
 & WorldMirror~\cite{liu2025worldmirror} & 1263M & 23.54 & 0.789 & 0.193 \\
 & ReSplat-Small (init)  &  62M & 26.77 & 0.865 & 0.142 \\
 & ReSplat-Base (init)   & 209M & 27.37 & 0.877 & 0.130 \\
 & ReSplat-Large (init)  & 559M & 27.86 & 0.886 & 0.121 \\
\midrule
\multirow{4}{*}{Recurrent}
 & ReSplat-Small (recurrent 1) & 77M & 28.17 & 0.890 & 0.118 \\
 & ReSplat-Small (recurrent 2) & 77M & 28.73 & 0.898 & 0.110 \\
 & ReSplat-Small (recurrent 3) & 77M & 28.96 & 0.901 & 0.107 \\
 & ReSplat-Small (recurrent 4) & 77M & \textbf{29.07} & \textbf{0.902} & \textbf{0.105} \\
\bottomrule
\end{tabular}
\label{tab:recurrent_vs_better_init}
\end{center}
\vspace{-8mm}
\end{table}

\boldstartspace{Generalization Across Datasets, View Counts, and Image Resolutions.} We evaluate the generalization capability of our model, which is trained exclusively on DL3DV at $512 \times 960$ resolution with 8 input views. First, when generalizing to the unseen RealEstate10K dataset (\cref{fig:cross_data_gen}), the improvement yielded by our recurrent model is more significant since, unlike single-step feed-forward models (iteration 0), our model adapts to the test data via rendering error, thus mitigating the domain gap. Second, we evaluate our initial and recurrent models with varying input view counts in \cref{fig:cross_view_gen}, and observe that our recurrent model benefits more from the additional input views, while the initial model saturates. This indicates that our rendering error-informed recurrent model exploits the additional information more effectively. Third, existing single-step feed-forward models usually exhibit significant performance degradation when the testing image resolution deviates from training. However, our recurrent model significantly improves the robustness to different testing resolutions (\cref{fig:cross_res_gen}). For example, our recurrent model improves by 5dB PSNR when generalizing from $512\times 960$ to $320 \times 640$. These experiments demonstrate that our recurrent model effectively adapts to out-of-distribution scenarios using the rendering error as a feedback signal, thus substantially enhancing robustness.

\boldstartspace{Single-Step \vs Recurrent Models.} In \cref{tab:recurrent_vs_better_init}, we compare our recurrent model with significantly larger single-step baselines. Our recurrent ReSplat-Small model comprises only 77M parameters, just 15M beyond its single-step initialization counterpart, yet it consistently surpasses all single-step baselines regardless of scale. Even with a single refinement iteration, ReSplat-Small (recurrent 1) achieves 28.17 dB PSNR, outperforming ReSplat-Large (init) which has 559M parameters and is 7$\times$ larger. After four iterations, the gap widens to 1.21 dB PSNR over ReSplat-Large (init) and 5.53 dB over WorldMirror~\cite{liu2025worldmirror}, which is $16\times$ larger with 1263M parameters. This demonstrates that recurrent refinement is fundamentally more parameter efficient than scaling a single-step model: the gains from iterative error correction cannot be matched by simply increasing model capacity. Rather than committing to a single feed-forward prediction, our model progressively corrects its estimates via the rendering-error feedback loop, allowing a compact network to surpass much larger single-step counterparts.

\begin{figure}[!t]
    \centering
    \begin{minipage}[t]{0.49\linewidth}
        \vspace{-2mm}
        \centering
        \includegraphics[width=0.9\linewidth]{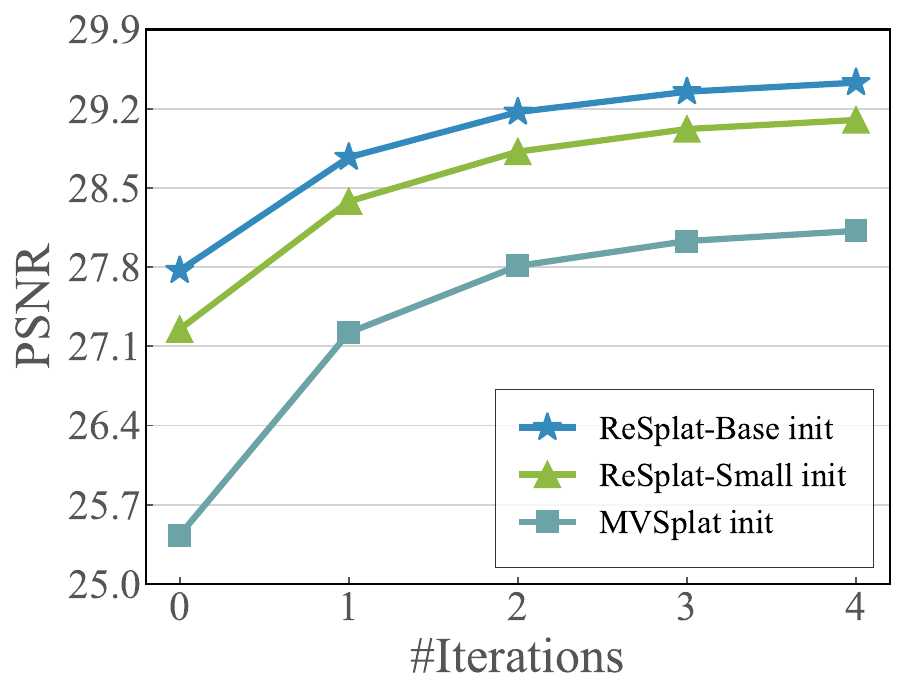}
        \vspace{-3mm}
        \caption{\textbf{ReSplat consistently improves across different initializations}.
        }
        \label{fig:diff_inits}
    \end{minipage}
    \hfill
    \begin{minipage}[t]{0.47\linewidth}
        \vspace{-1mm}
        \centering
        \includegraphics[width=\linewidth]{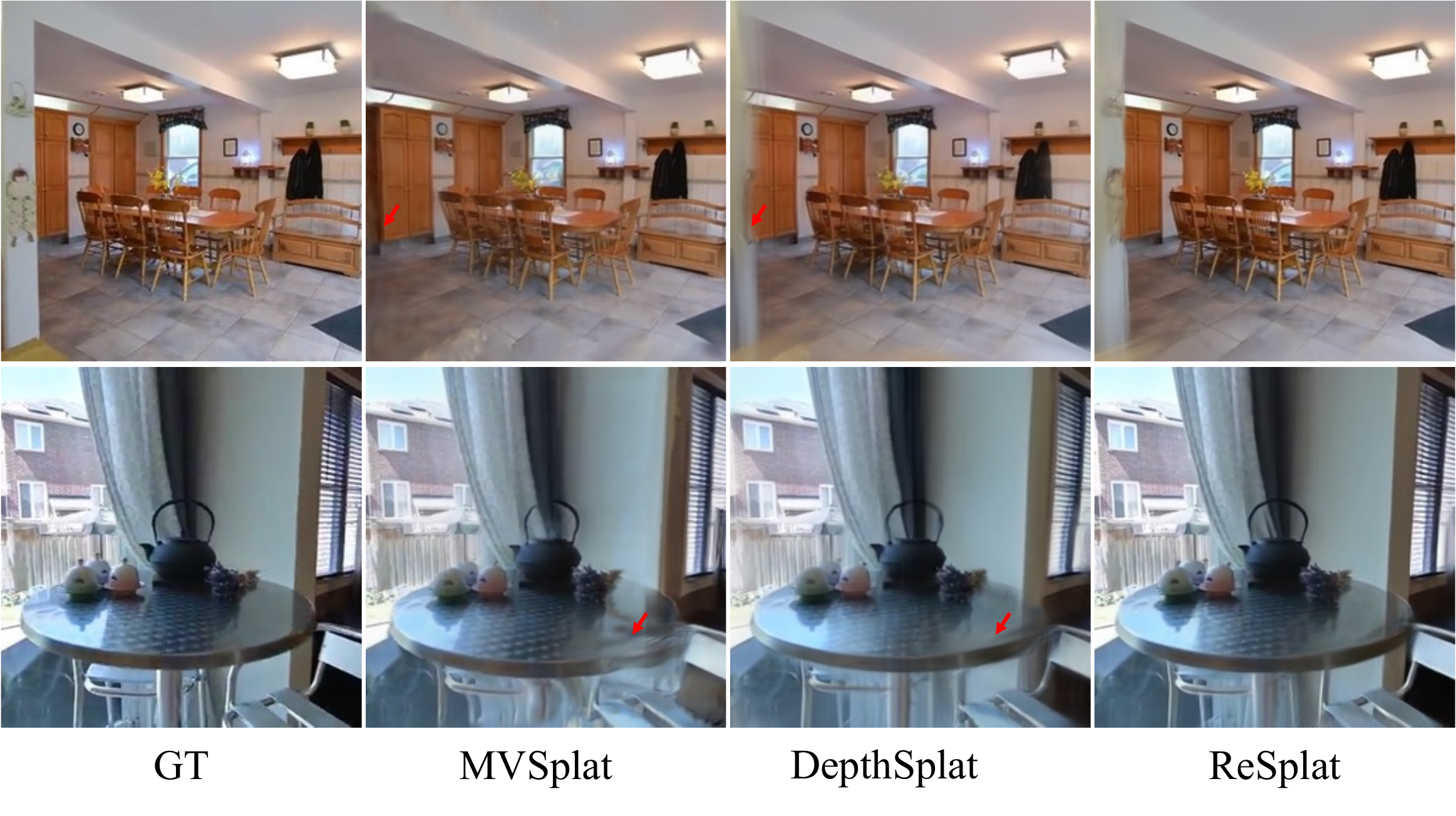}
        \vspace{-5mm}
        \caption{\textbf{Visual comparisons on RealEstate10K}. 
        ReSplat produces sharper structures than MVSplat and DepthSplat.}
        \label{fig:vis_re10k}
    \end{minipage}
    \vspace{-2mm}
\end{figure}

\boldstartspace{Different Initializations}. As shown in \cref{fig:diff_inits}, our recurrent model consistently improves rendering quality across various initializations: MVSplat~\cite{chen2024mvsplat}, ReSplat-Small, and ReSplat-Base. Performance improves monotonically over successive iterations regardless of the starting point, with stronger initializations yielding higher final quality. Furthermore, because MVSplat predicts per-pixel Gaussians, it produces $16\times$ more Gaussians than ReSplat, making subsequent refinement $13\times$ slower. In contrast, our compact initialization simultaneously provides a superior starting point and enables highly efficient recurrent updates.

\begin{table}[!t]

\begin{center}
\footnotesize

\caption{\textbf{Evaluation of 16 input views ($540 \times 960$) on DL3DV.} Our ReSplat reconstructs $4\times$ fewer Gaussians than Long-LRM but still outperforms it.
    }
    \vspace{-2mm}
    \setlength{\tabcolsep}{1.5pt} %
    \begin{tabular}{lcccccccccccccccccccccccc}
    \toprule
    Method & \#Iterations & PSNR $\uparrow$ & SSIM $\uparrow$ & LPIPS $\downarrow$ & Recon. Time & \#Gaussians & \\

    \midrule

    3DGS~\cite{Kerbl2023TOG} & 30000 & 21.20 & 0.708 & 0.264 & 13min & - \\
    Mip-Splatting~\cite{yu2024mipsplatting} & 30000 & 20.88 & 0.712 & 0.274 & 13min & - \\
    Scaffold-GS~\cite{lu2024scaffold} & 30000 & 22.13 & 0.738 & \textbf{0.250} & 16min & - \\
    Long-LRM~\cite{chen2025llrm} & 0 & 22.66 & 0.740 & 0.292 & \textbf{0.4sec} & 2073K \\
    \multirow{3}{*}{ReSplat} & 0 & 22.69 & 0.742 & 0.307 & 0.7sec & \textbf{518K}\\
     & 1 & 23.23 & 0.758 & 0.291 & 1.2sec &  \textbf{518K} \\
     & 2 & \textbf{23.51} & \textbf{0.766} & 0.284 & 1.7sec &  \textbf{518K} \\
    
    \bottomrule
    \end{tabular}
    \label{tab:16view_high_dl3dv}
    \end{center}
    \vspace{-8mm}
\end{table}

\boldstartspace{16 Views at $540 \times 960$ Resolution.} We follow Long-LRM~\cite{chen2025llrm} for this evaluation setup such that a direct comparison is possible. The results of 3DGS~\cite{Kerbl2023TOG}, Mip-Splatting~\cite{yu2024mipsplatting}, and Scaffold-GS~\cite{lu2024scaffold} are borrowed from Long-LRM paper. This experiment aims to reconstruct the full DL3DV scene from 16 input views, which is particularly challenging due to the expansive spatial coverage of the DL3DV dataset. However, our ReSplat still outperforms previous optimization and feed-forward methods, as shown in \cref{tab:16view_high_dl3dv}. Notably, Long-LRM uses Gaussian pruning based on opacity during training and evaluation, resulting in a $\sim 4 \times$ reduction in the number of Gaussians. In contrast, we compress the Gaussians by $16\times$, thus our final reconstruction has $4\times$ fewer Gaussians than Long-LRM while still outperforming it. Our reconstruction time is slower than Long-LRM, mainly due to the $k$NN operation. Further implementation-level optimizations could potentially improve our reconstruction speed.

\begin{figure}[!t]
    \centering
    \begin{minipage}[t]{0.48\linewidth}
        \centering
        \setlength{\tabcolsep}{2pt}
        \captionof{table}{\textbf{Evaluation of two input views on RealEstate10K}. 
        }
        \resizebox{\textwidth}{!}{\begin{tabular}{lcccc}
        \toprule
        Method & w/ 3DGS & PSNR $\uparrow$ & SSIM $\uparrow$ & LPIPS $\downarrow$ \\
        \midrule
        pixelSplat~\cite{charatan2023pixelsplat}     & \ding{51} & 25.89 & 0.858 & 0.142 \\
        MVSplat~\cite{chen2024mvsplat}        & \ding{51} & 26.39 & 0.869 & 0.128 \\
        DepthSplat~\cite{xu2025depthsplat}     & \ding{51} & 27.47 & 0.889 & 0.114 \\
        GS-LRM~\cite{zhang2024gs}         & \ding{51} & 28.10 & 0.892 & 0.114 \\
        Long-LRM~\cite{chen2025llrm}       & \ding{51} & 28.54 & 0.895 & 0.109 \\
        LVSM (enc-dec)~\cite{jin2024lvsm}          & \ding{55} & 28.58 & 0.893 & 0.114 \\
        LVSM (dec-only)~\cite{jin2024lvsm}        & \ding{55} & 29.67 & 0.906 & \textbf{0.098} \\
        ReSplat & \ding{51} & \textbf{29.75} & \textbf{0.912} & 0.100 \\
        \bottomrule
        \end{tabular}}
        \label{tab:sota_re10k}
    \end{minipage}
    \hfill
    \begin{minipage}[t]{0.48\linewidth}
        \centering
        \captionof{table}{\textbf{RealEstate10K to ACID cross-dataset generalization}. 
        }
        \label{tab:acid_genelization}
        
        \begin{tabular}{lccc}
            \toprule
            Method & PSNR $\uparrow$ & SSIM $\uparrow$ & LPIPS $\downarrow$ \\
            \midrule
            pixelSplat~\cite{charatan2023pixelsplat} & 27.64 & 0.830 & 0.160 \\
            MVSplat~\cite{chen2024mvsplat} & 28.15 & 0.841 & 0.147 \\
            DepthSplat~\cite{xu2025depthsplat} & 28.37 & 0.847 & 0.141 \\
            GS-LRM~\cite{zhang2024gs} & 28.84 & 0.849 & 0.146 \\
            ReSplat & \textbf{29.87} & \textbf{0.864} & \textbf{0.135} \\
            \bottomrule
        \end{tabular}
    \end{minipage}
    \vspace{-2mm}
\end{figure}

\vspace{1pt}
\boldstartspace{2 Views at $256 \times 256$ Resolution}. Since redundancy is less prevalent in two-view, low-resolution ($256 \times 256$) scenarios, we employ $4\times$ spatial subsampling in the 3D space and decode 4 Gaussians from each subsampled 3D point in our initial reconstruction model. Consequently, the total number of Gaussians remains consistent with previous per-pixel methods. The recurrent process remains the same as the many-view setups. 
Table~\ref{tab:sota_re10k} shows that our ReSplat outperforms previous feed-forward 3DGS models (\eg, DepthSplat~\cite{xu2025depthsplat}, GS-LRM~\cite{zhang2024gs} and Long-LRM~\cite{chen2025llrm}) by significant margins. 
Compared to the 3DGS-free method LVSM~\cite{jin2024lvsm}, we outperform its encoder-decoder architecture by 1.1dB PSNR, and our results are competitive with its best-performing decoder-only model variant. However, our method offers the benefits of an explicit 3D Gaussian representation, enabling a $20\times$ increase in rendering speed. We present visual comparisons in \cref{fig:vis_re10k}, where our ReSplat produces sharper structures than MVSplat and DepthSplat. In \cref{tab:acid_genelization}, we show the zero-shot generalization results on the unseen ACID~\cite{liu2021infinite} dataset, where our ReSplat again outperforms previous methods by clear margins.

\boldstartspace{Comparison with SplatFormer.} We note that our ReSplat has several crucial differences with SplatFormer~\cite{chen2024splatformer}. First, we identify the rendering error as an informative feedback signal for improving Gaussian reconstruction, which is missing in SplatFormer. Second, ReSplat is a recurrent model which supports multi-step refinement with a weight-sharing architecture, while SplatFormer is a non-recurrent network designed for single-step refinement. Third, ReSplat is a pure feed-forward model with feed-forward initialization and feed-forward refinement. In contrast, SplatFormer relies on lengthy optimization-based 3DGS to get initial Gaussians. Fourth, SplatFormer is designed for object-centric datasets, while ReSplat can handle diverse scene-level datasets where the complexity is much higher than objects. Despite these differences, we tried to conduct a comparison with SplatFormer on our scene-level datasets. We found it particularly challenging to make it work properly for scene-level datasets, since it relies on Point Transformer V3~\cite{wu2024ptv3} where a proper grid size is required to serialize the point cloud. This can be done for object-centric datasets where normalizing the objects to $[-1, 1]$ is possible. However, for unbounded scene-level datasets, this would be very challenging. We tried different normalizations and did grid search for the grid size, and the best results we obtained with SplatFormer are clearly worse than ReSplat across all metrics (ReSplat \vs SplatFormer, PSNR: 29.07 \vs 27.03, SSIM: 0.902 \vs 0.868, LPIPS: 0.105 \vs 0.140).

In the appendix, we provide additional comparisons with optimization-based 3DGS~\cite{Kerbl2023TOG} across 8, 16, and 32 views, and depth-regularized sparse-view 3DGS optimization methods~\cite{chung2024depth,li2024dngaussian}.

\vspace{-4pt}
\subsection{Analysis and Ablation}
\vspace{-4pt}

We conduct several experiments to analyze the behavior of our architecture and validate our design choices.
To save compute, all experiments in this section are performed using 8 input views at $256 \times 448$ resolution on the DL3DV dataset.

\begin{table}[t]
\centering
\caption{\textbf{Ablations}.}
\vspace{-16pt}
\captionsetup[subfloat]{position=top}
\subfloat[\textbf{Ablation of the rendering error}.\label{tab:ablation_render_error}]{
\begin{minipage}[t]{0.52\linewidth}
\centering
\small
\setlength{\tabcolsep}{1.5pt} 
\resizebox{\linewidth}{!}{
\begin{tabular}{lccccc}
\toprule
Method & PSNR ↑ & SSIM ↑ & LPIPS ↓ \\
\midrule
Initialization & 26.77 & 0.865 & 0.142 \\
\midrule
w/o rendering error & 27.19 & 0.873 & 0.137 \\
RGB error only & 27.90 & 0.882 & 0.130 \\
Feature error only & 28.77 & 0.897 & 0.110 \\
Concat (RGB \& feature errors) & 28.93 & 0.900 & 0.106 \\
Add (RGB \& feature errors) & \textbf{29.07} & \textbf{0.902} & \textbf{0.105} \\
\bottomrule
\end{tabular}
}
\end{minipage}
}
\hfill
\subfloat[\textbf{Ablation of the coordinate system}.\label{tab:ablation_coord_system}]{
\begin{minipage}[t]{0.42\linewidth}
\centering
\small
\setlength{\tabcolsep}{1.5pt} 
\resizebox{0.9\linewidth}{!}{
\begin{tabular}{lccccc}
\toprule
Method & PSNR ↑ & SSIM ↑ & LPIPS ↓ \\
\midrule
Initialization & 26.77 & 0.865 & 0.142 \\
\midrule
COLMAP & 28.14 & 0.886 & 0.116 \\
First view & 28.66 & 0.896 & 0.109 \\
Last view & 28.59 & 0.895 & 0.110 \\
Middle view & \textbf{29.07} & \textbf{0.902} & \textbf{0.105} \\
\bottomrule
\end{tabular}
}
\end{minipage}
}

\vspace{-2pt}

\subfloat[\textbf{Ablation of the initial reconstruction model}.\label{tab:ablation_init}]{
\begin{minipage}[t]{0.52\linewidth}
\centering
\small
\setlength{\tabcolsep}{1.5pt} 
\resizebox{\linewidth}{!}{
\begin{tabular}{lccccc}
\toprule
Method & PSNR ↑ & SSIM ↑ & LPIPS ↓ & \#Gaussians \\
\midrule
DepthSplat~\cite{xu2025depthsplat} & 25.79 & 0.861 & \textbf{0.134} & 918K \\
\midrule
Initialization   & \textbf{26.77} & \textbf{0.865} & 0.142  & \textbf{57K} & \\
w/o $k$NN attn  & 25.30 &  0.833 &   0.178    & \textbf{57K} \\
w/o global attn  & 26.33  & 0.856 & 0.150  & \textbf{57K} \\
w/o $k$NN, w/o global  & 24.50 & 0.814 & 0.200 & \textbf{57K} \\
\bottomrule
\end{tabular}
}
\end{minipage}
}
\hfill
\subfloat[\textbf{Ablation of the recurrent model}.\label{tab:ablation_refine}]{
\begin{minipage}[t]{0.42\linewidth}
\centering
\small
\setlength{\tabcolsep}{1.5pt} 
\resizebox{0.9\linewidth}{!}{
\begin{tabular}{lccccc}
\toprule
Method & PSNR ↑ & SSIM ↑ & LPIPS ↓ \\
\midrule
Initialization & 26.77 & 0.865 & 0.142 \\
\midrule
Full & \textbf{29.07} & \textbf{0.902} & \textbf{0.105} \\
w/o state & 27.79 & 0.878 & 0.125 \\
w/o $k$NN attn & 28.58 & 0.894 & 0.111 \\
w/o global attn & 28.96 & 0.900 & 0.107 \\
\bottomrule
\end{tabular}
}
\end{minipage}
}

\vspace{-10pt}
\end{table}

\boldstartspace{Rendering Error.} We evaluate the impact of the rendering error in \cref{tab:ablation_render_error}. Removing the rendering error in our recurrent model results in a significant performance drop (-1.9dB PSNR). The feature-space errors are more effective than the pixel-space errors, and the best performance is obtained by combining both (addition is slightly better than concatenation).

\boldstartspace{Coordinate System.} In \cref{tab:ablation_coord_system}, we observe that aligning to the middle input view's camera pose performs significantly better (+0.9dB PSNR) than using the default global coordinate system provided by COLMAP. We attribute this to the spatial distribution of the views, where anchoring to the central view acts as a pivot, balancing the spatial distribution of the 3D Gaussians and facilitating the learning of 3D spatial relationships.

\boldstartspace{Ablation of the Initial Model.} As shown in \cref{tab:ablation_init}, $k$NN attention is crucial for maintaining performance when compressing the Gaussian count by $16\times$. Global attention also yields moderate gains, indicating that both local and global 3D contexts are essential for learning compact 3D representations. 
Together, these components enable our initial model to outperform DepthSplat, despite using $16\times$ fewer Gaussians. The visual results are in the appendix.

\boldstartspace{Ablation of the Recurrent Model.} The state input (\cref{eq:init_gaussian}) is critical to our recurrent network (\cref{tab:ablation_refine}). Unlike the raw, low-level Gaussian attributes, it encodes rich latent features derived from our initialization model. Both $k$NN attention and global attention contribute to the performance. Corresponding visual ablations are provided in the appendix.

In the appendix, we provide further evaluations of different feature types for computing the rendering error, recurrent \vs non-recurrent architectures, as well as different compression factors ($4\times$, $16\times$, and $64 \times$) in the initialization model, model profiling, and qualitative results across varying iteration counts.

\section{Conclusion}

We presented ReSplat, a feed-forward recurrent Gaussian splatting model that enables efficient and high-quality view synthesis. By leveraging the rendering error as a feedback signal and operating in a compact subsampled 3D space, our method significantly reduces the number of Gaussians while improving performance and generalization across datasets, view counts, and resolutions.

\boldstartspace{Limitations}. Our current model maintains a fixed Gaussian count during refinement. Integrating adaptive pruning and densification strategies~\cite{Kerbl2023TOG} could potentially further improve performance. In addition, ReSplat currently saturates after four iterations. We hypothesize that the recurrent network rapidly extracts the most informative rendering error signals during the earliest few steps, and performance gradually plateaus due to the diminishing returns of local refinement. Exploring more informative feedback mechanisms to effectively scale test-time compute remains a promising future direction. Furthermore, ReSplat currently operates on static view sets; it would be interesting to extend it to streaming scenarios using a sliding-window approach. In this setup, Gaussians could be initialized independently per window, and the recurrent module could continuously refine the scene, either locally or globally, as new frames arrive. We view these directions as promising avenues for future research.

\boldstartspace{Acknowledgments}. We thank Naama Pearl, Xudong Jiang, Stefano Esposito, and Ata Celen for the insightful comments, and Yung-Hsu Yang and Kashyap Chitta for the fruitful discussions. We thank the anonymous reviewers for their constructive feedback. Andreas Geiger was supported by the ERC Starting Grant LEGO-3D (850533) and the DFG EXC number 2064/1 - project number 390727645. This work was supported as part of the Swiss AI Initiative by a grant from the Swiss National Supercomputing Centre (CSCS) under project ID a144 on Alps.

\bibliographystyle{splncs04}
\bibliography{main}

\newpage

\setcounter{section}{0}
\setcounter{figure}{0}
\setcounter{table}{0}

\renewcommand{\thesection}{\Alph{section}} 

\renewcommand{\thefigure}{S\arabic{figure}}
\renewcommand{\thetable}{S\arabic{table}}

\section*{Appendix}

This appendix provides further details and comprehensive evaluations to complement the main manuscript. In \cref{sec:supp_eval}, we present extended evaluations, including comparisons with optimization-based 3DGS~\cite{Kerbl2023TOG} across varying input views and depth-regularized 3DGS~\cite{chung2024depth,li2024dngaussian}. We also provide detailed ablations on our recurrent architecture, feature extraction choices, compression factors, and our efficient local $k$NN implementation, alongside comprehensive model profiling. In \cref{sec:supp_details}, we detail our specific training schedules and hardware configurations. In \cref{sec:supp_viz}, we supply extensive additional qualitative results, featuring visual comparisons with state-of-the-art baselines, and progressive refinement visualizations across iterations.

\section{Additional Evaluations}
\label{sec:supp_eval}

\boldstartspace{Comparison with 3DGS Across 8, 16, and 32 Views.} In \cref{tab:resplat_vs_3dgs}, we compare with optimization-based 3DGS~\cite{Kerbl2023TOG} across 8, 16 and 32 input views. The quality gap becomes smaller when given 32 views for optimization-based 3DGS. However, our ReSplat is more than $200 \times$ faster in terms of the reconstruction speed. In this setup, we expand the sampling region as the number of input views increases to enlarge scene coverage. Consequently, the test views differ across configurations to account for this larger spatial extent.

\begin{table}[h]

\begin{center}
\footnotesize

\caption{\textbf{Comparison with optimization-based 3DGS across 8, 16, and 32 input views.} The quality gap becomes smaller when given 32 views for optimization-based 3DGS. However, our ReSplat is more than $200\times$ faster in terms of the reconstruction speed. The image resolution is $256 \times 448$. 
    }
    \vspace{-8pt}
    \resizebox{\linewidth}{!}{
    \begin{tabular}{ccccccccccccccccccccccccc}
    \toprule
    \#Views & Method & Category & \#Iterations & PSNR $\uparrow$ & SSIM $\uparrow$ & LPIPS $\downarrow$ & \#Gaussians & {\begin{tabular}[x]{@{}c@{}}Recon.\\Time (s) \end{tabular}} \\
    
    \midrule
    \multirow{2}{*}{8} & 3DGS & Optimization & 4000 & 26.44 & 0.841 & 0.134 & 250K & 49 \\
    & ReSplat & Feed-Forward & \textbf{4} & \textbf{29.20} & \textbf{0.904} & \textbf{0.104} & \textbf{57K} & \textbf{0.21} \\

    \midrule
    \multirow{2}{*}{16} & 3DGS & Optimization & 4000 & 27.38 & 0.864 & 0.119 & 395K & 70 \\
    & ReSplat & Feed-Forward & \textbf{4} &\textbf{ 29.01} & \textbf{0.900} & \textbf{0.105} & \textbf{114K} & \textbf{0.34} \\

    \midrule
    \multirow{2}{*}{32} & 3DGS & Optimization & 4000 & 27.86 & 0.879 & \textbf{0.113} & 522K  & 160 \\
    & ReSplat & Feed-Forward & \textbf{4} & \textbf{28.30} & \textbf{0.891} & 0.114 & \textbf{229K} & \textbf{0.75} \\

    \bottomrule
    \end{tabular}
    }
    
    \label{tab:resplat_vs_3dgs}
    \end{center}
    \vspace{-8pt}
\end{table}

\boldstartspace{Recurrent \vs Non-recurrent Architecture.} We demonstrate the effectiveness of our recurrent model by comparing with non-recurrent variants in \cref{tab:recurrent}. In particular, we first compare with non-weight-sharing multi-step stacked networks where different iterations have different model weights and all the other components are the same. We can observe that non-weight-sharing not only leads to $4 \times$ more parameters, but also results in worse view synthesis results. We further compare with non-weight-sharing single-step deeper networks by increasing the number of attention blocks for a single-step refinement, where the results are clearly worse than our multi-step recurrent network. The weight-sharing design in our recurrent network implicitly regularizes training, which is not only more parameter-efficient but also leads to better results.

\begin{table}[!h]
    \centering
    \caption{\textbf{Comparison of recurrent and non-recurrent architectures.} Compared to non-weight-sharing multi-step stacked networks and non-weight-sharing single-step deeper networks, our weight-sharing multi-step recurrent network is not only more parameter-efficient, but also leads to better results. We also note that our single recurrent network can support different numbers of iterations thanks to weight-sharing.}
    \vspace{-8pt}
    \label{tab:recurrent}
    \begin{tabular}{lcccc}
        \toprule
        Configuration & \#Params & PSNR $\uparrow$ & SSIM $\uparrow$ & LPIPS $\downarrow$ \\
        \midrule
        \textit{weight-sharing} & & & & \\
        Recurrent (iter 1, block 4) & 13.8M & 28.17 & 0.890 & 0.118 \\
        Recurrent (iter 2, block 4) & 13.8M & 28.73 & 0.898 & 0.110 \\
        Recurrent (iter 3, block 4) & 13.8M & 28.96 & 0.901 & 0.107 \\
        Recurrent (iter 4, block 4) & 13.8M & \textbf{29.07} & \textbf{0.902} & \textbf{0.105} \\
        \midrule
        \textit{non-weight-sharing, multi-step, stacked} & & & & \\
        Non-recurrent (stack 1)     & 13.8M & 28.17 & 0.890 & 0.118 \\
        Non-recurrent (stack 2)     & 27.6M & 28.74 & 0.898 & 0.109 \\
        Non-recurrent (stack 3)     & 41.4M & 28.72 & 0.898 & 0.110 \\
        Non-recurrent (stack 4)     & 55.2M & 28.71 & 0.897 & 0.110 \\
        \midrule
        \textit{non-weight-sharing, single-step, deeper} & & & & \\
        Non-recurrent  (stack 1, block 4) & 13.8M & 28.17 & 0.890 & 0.118 \\
        Non-recurrent (stack 1, block 8)  & 27.6M & 28.30 & 0.891 & 0.116 \\
        Non-recurrent (stack 1, block 12) & 41.4M & 28.36 & 0.893 & 0.115 \\
        Non-recurrent (stack 1, block 16) & 55.2M & 28.40 & 0.893 & 0.115 \\
        \bottomrule
    \end{tabular}
    \vspace{-8pt}
\end{table}

\boldstartspace{Comparison with Sparse-View Optimization Methods.} We additionally compare with optimization-based 3DGS methods that are specifically designed for sparse input views. These sparse-view optimization methods~\cite{li2024dngaussian,chung2024depth} usually rely on additional depth losses to regularize the optimization process. To compare with them, we perform 3DGS optimization with an additional depth loss between the rendered depth map and the estimated monocular depth map from Depth Anything V2 (Large)~\cite{yang2024depth}. The results are reported in \cref{tab:3dgs_with_depth_loss}. With the additional depth loss, the 3DGS optimization results are improved by 1dB PSNR. However, the gap with our ReSplat is still significant (3dB PSNR). The additionally introduced depth loss also makes the optimization slower due to the additional time for depth rendering and monocular depth estimation. In contrast, our model doesn't rely on any additional supervision from an external monocular depth model and it's $94\times$ faster thanks to our feed-forward nature.

\begin{table}[h]
    \centering
    \caption{\textbf{Comparison with depth-regularized 3DGS optimization method.} Despite with an additional depth loss, the optimization-based method is still 3dB PSNR worse than our ReSplat and runs $94\times $ slower.}
    \label{tab:3dgs_with_depth_loss}
    \vspace{-8pt}
    \begin{tabular}{lcccc}
        \toprule
        Method &  PSNR $\uparrow$ & SSIM $\uparrow$ & LPIPS $\downarrow$ & Recon. Time (s) \\
        \midrule
        3DGS (w/o depth loss) & 23.46 & 0.770 & 0.224 & 70.0 \\
        3DGS (w/ depth loss) & 24.54 & 0.796 & 0.204 & 75.4 \\
        ReSplat (w/o depth loss) & \textbf{27.70} & \textbf{0.868 }& \textbf{0.160} & \textbf{0.8} \\
        \bottomrule
    \end{tabular}
    \vspace{-8pt}
\end{table}

\boldstartspace{Features for Computing the Rendering Error.} In \cref{tab:feature_for_render_error}, we compare ResNet~\cite{he2016deep} features with those from DINOv2~\cite{oquab2023dinov2}. We observed no improvement when using the larger, more recent feature extractor. We attribute this to the patch-based architecture of DINOv2, which may result in coarser spatial information. In contrast, convolutional networks maintain local structural fidelity, which is critical for high-quality pixel-accurate view synthesis.

\begin{table}[t]
    \centering
    \caption{\textbf{ResNet \vs DINOv2 features for computing the rendering error.} We observed no improvement when using the larger, more recent feature extractor. We attribute this to the patch-based architecture of DINOv2, which may result in coarser spatial information. In contrast, convolutional networks maintain local structural fidelity, which is critical for high-quality view synthesis.}
    \label{tab:feature_for_render_error}
    \vspace{-6pt}
    \begin{tabular}{lcccc}
        \toprule
        Features & \#Parameters &  PSNR $\uparrow$ & SSIM $\uparrow$ & LPIPS $\downarrow$ \\
        \midrule
        ResNet & 0.7M & 29.07 & 0.902 & 0.105 \\
        DINOv2 & 86.6M & 29.00 & 0.901 & 0.107 \\
        \bottomrule
    \end{tabular}
    \vspace{-8pt}
\end{table}

\begin{figure}[!h]
    \centering
    \includegraphics[width=\linewidth]{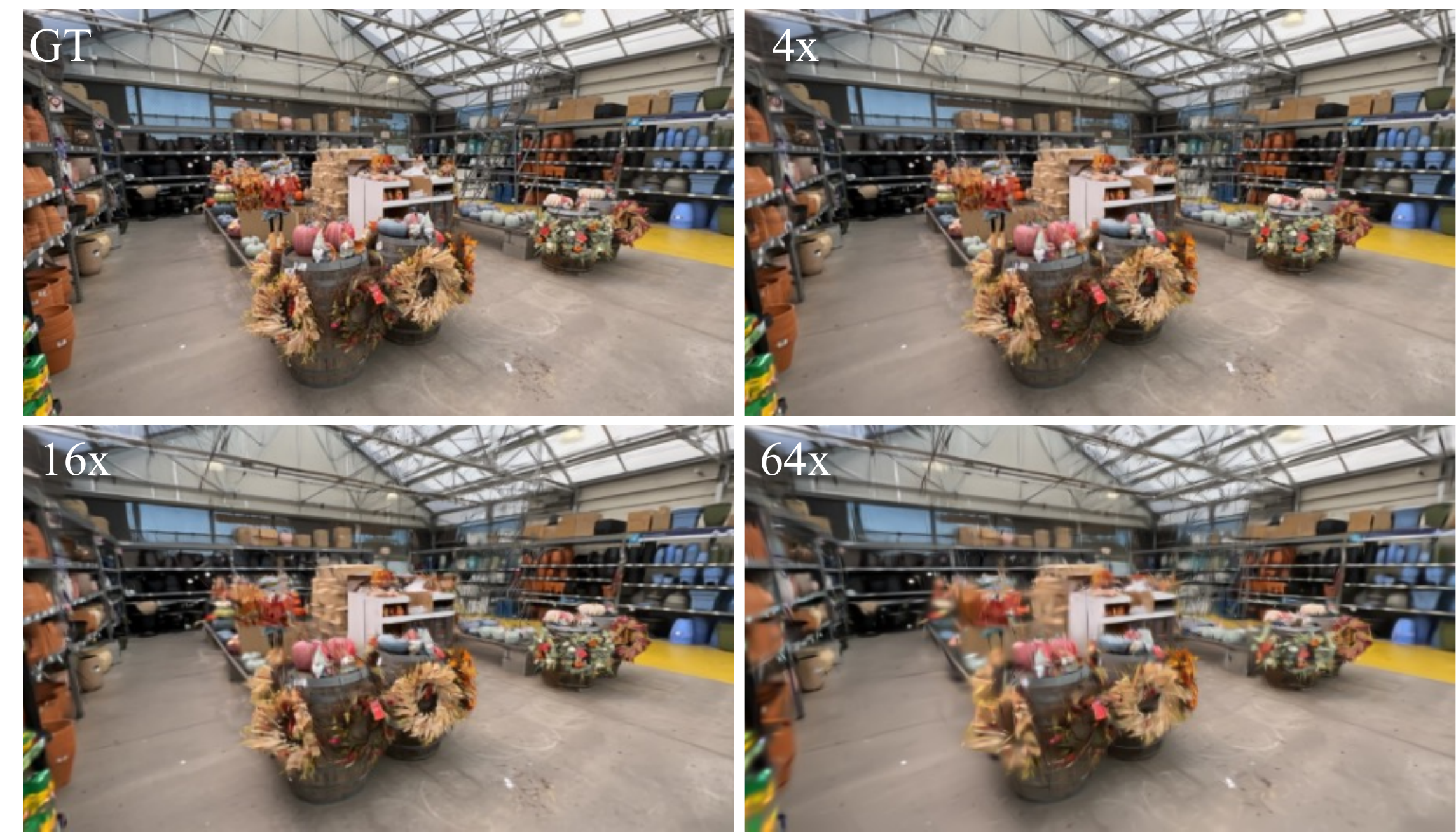}
    \vspace{-4mm}
    \caption{\textbf{Different compression factors.} $16\times$ compression represents a good speed-accuracy trade-off. 
    }
    \label{fig:compression_supp}
    \vspace{-4mm}
\end{figure}

\boldstartspace{Compression Factor.} By default, we compress the number of Gaussians by $16\times$ using depth maps at $1/4$ resolution. In \cref{tab:compression_factor} and \cref{fig:compression_supp}, we compare with $64 \times$ (with $1/8$ depth maps) and $4 \times$ (with $1/2$ depth maps) compression factors. We observe that less compression leads to higher quality. However, $4 \times$ compression is $2\times$ slower than $16 \times$ at $256 \times 448$ resolution. It would be more expensive when handling higher resolution images (\eg, $512 \times 960$). Thus, we choose $16 \times$ compression as a good speed-accuracy trade-off.

\begin{table}[!h]
    \centering
    \vspace{-10pt}
    \caption{\textbf{Different compression factors.} $16\times$ compression represents a good speed-accuracy trade-off and thus is used in our model.}
    \label{tab:compression_factor}
    \begin{tabular}{ccccc}
        \toprule
            Compression & PSNR $\uparrow$ & SSIM $\uparrow$ & LPIPS $\downarrow$ & Time (s) \\
            \midrule
            $64 \times$ & 24.77 & 0.797 & 0.226 & \textbf{0.096} \\
            $16 \times$ & 26.77 & 0.865 & 0.142 & 0.104 \\
            $4 \times$ & \textbf{28.36} & \textbf{0.900} & \textbf{0.103} & 0.206 \\
        \bottomrule
    \end{tabular}
    \vspace{-25pt}
\end{table}

\boldstartspace{Efficient Local $k$NN Implementation.} The standard $k$NN attention implementation~\cite{pointcept2023} used in this paper relies on a global brute-force search over all $N$ points, imposing an $O(N^2)$ complexity bottleneck for high-resolution point cloud. To address this, we introduce a drop-in $O(N)$ local $k$NN module that exploits known multi-view grid structure. Because nearby pixels on a surface project to nearby 3D points, the true $k$-nearest neighbors are almost exclusively found at spatially adjacent pixels in the source view, or at corresponding projected pixels in other views. We therefore constrain our search spatially while remaining comprehensive across the multi-view dimension. For each point, we generate same-view candidates using a $(2r_s+1)^2$ spatial window. Simultaneously, we gather cross-view candidates by projecting the 3D query point into the 2D image planes of all other available cameras.  Specifically, we use each target camera's pose to locate the point relative to that camera, and its camera intrinsic matrix to pinpoint the exact 2D pixel where the point would be visible. We then extract the 3D points located within a $(2r_c+1)^2$ spatial window centered around this projected pixel coordinate. 
For our implementation, we set both spatial radii $r_s$ and $r_c$ to 3. Second, we select the top-$k$ neighbors from these candidates using 3D Euclidean distance. By vectorizing the cross-view reprojections (with camera intrinsic and extrinsic parameters), our method reduces complexity to $O(N)$, significantly reducing distance computations with negligible quality loss. As demonstrated in \cref{tab:local_knn}, our local $k$NN implementation improves inference speed over the global baseline when evaluated on 8 views at $512 \times 960$ resolution, empirically validating its computational efficiency.

\begin{table}[!h]
    \centering
    \caption{\textbf{Global \vs Local $k$NN.} Our local $k$NN implementation improves inference speed over the default global baseline. The results are evaluated on 8 views at $512 \times 960$ resolution (with 246K Gaussians).}
    \vspace{-10pt}
    \label{tab:local_knn}
    \begin{tabular}{ccccc}
        \toprule
            $k$NN & PSNR $\uparrow$ & SSIM $\uparrow$ & LPIPS $\downarrow$ & Time (s) \\
            \midrule
            Global & \textbf{27.70} & \textbf{0.868} & \textbf{0.160} & 0.816 \\
            Local & 27.65 & 0.867 & 0.163 & \textbf{0.591} \\
        \bottomrule
    \end{tabular}
    \vspace{-10pt}
\end{table}

\boldstartspace{Model Profiling.} In \cref{tab:profiling}, we report the total runtime and individual component latency. In the initial reconstruction model, the depth prediction module constitutes the majority of the runtime. For the recurrent model, the kNN attention mechanism consumes the most time. These results highlight potential areas for future optimization.

\begin{table}[h]
    \centering
    \caption{\textbf{Model Profiling.} Inference time (s) measured on 8 input views at varying resolutions. We report both total runtime and individual component latency.} 
    \label{tab:profiling}
    
    \begin{subtable}[t]{0.48\textwidth}
        \centering
        \caption{\textbf{Initial Model Profiling.} The depth module constitutes the majority of the runtime.}
        \label{tab:init_profiling}
        \resizebox{\textwidth}{!}{%
        \begin{tabular}{lccccc}
            \toprule
            Resolution & Total & {\begin{tabular}[x]{@{}c@{}}Depth\\pred. \end{tabular}} & {\begin{tabular}[x]{@{}c@{}}$k$NN\\attn \end{tabular}} & {\begin{tabular}[x]{@{}c@{}}Global\\attn \end{tabular}} & {\begin{tabular}[x]{@{}c@{}}Gaussian\\head \end{tabular}} \\
            \midrule
            $256 \times 448$ & 0.149 & 0.111 & 0.024 & 0.013 & 0.001 \\
            $512 \times 960$ & 0.311 & 0.197 & 0.094 & 0.018 & 0.002 \\
            \bottomrule
        \end{tabular}%
        }
    \end{subtable}%
    \hfill
    \begin{subtable}[t]{0.48\textwidth}
        \centering
        \caption{\textbf{Recurrent Model Profiling.} The $k$NN attention mechanism consumes the most time.}
        \label{tab:recurrent_profiling}
        \resizebox{\textwidth}{!}{%
        \begin{tabular}{lccccc}
            \toprule
            Resolution & Total & {\begin{tabular}[x]{@{}c@{}}Render\\error \end{tabular}} & {\begin{tabular}[x]{@{}c@{}}$k$NN\\attn \end{tabular}} & {\begin{tabular}[x]{@{}c@{}}Global\\attn \end{tabular}} & {\begin{tabular}[x]{@{}c@{}}Update\\head \end{tabular}} \\
            
            \midrule
            $256 \times 448$ & 0.022 & 0.003 & 0.015 & 0.002 & 0.002 \\
            $512 \times 960$ & 0.126 & 0.016 & 0.092 & 0.008 & 0.010 \\
            \bottomrule
        \end{tabular}%
        }
    \end{subtable}
\end{table}

\boldstartspace{Per-Attribute Update.} Our per-attribute analysis shows that the recurrent module primarily refines geometry and base appearance (Gaussian centers, scales, and base colors ($0$-th order)), whereas higher-order SH adjustments remain negligible.

\section{Additional Details}
\label{sec:supp_details}

\boldstartspace{Training Details.} We train our model with cosine learning rate schedule. For experiments on DL3DV, we adopt a progressive training strategy with gradually increased image resolutions and number of input views for better efficiency. More specifically, we first train our model with 8 input views at $256 \times 448$ resolution, and then we fine-tune the model with 8 input views at $512 \times 960$ resolution, and finally we fine-tune the model with 16 input views at $512 \times 960$ resolution. For each stage, we train with 16 GH200 GPUs for 80K steps, with 50K steps for the initial reconstruction model and 30K steps for the recurrent model. For experiments on RealEstate10K at $256 \times 256$ resolution, we first train the initial model with 16 GH200 GPUs for 200K steps and then train the recurrent model for 100K steps. More details are provided at \url{https://github.com/cvg/resplat}.

\section{Additional Visualizations}
\label{sec:supp_viz}

\begin{figure}[!t]
    \centering
    \includegraphics[width=0.98\linewidth]{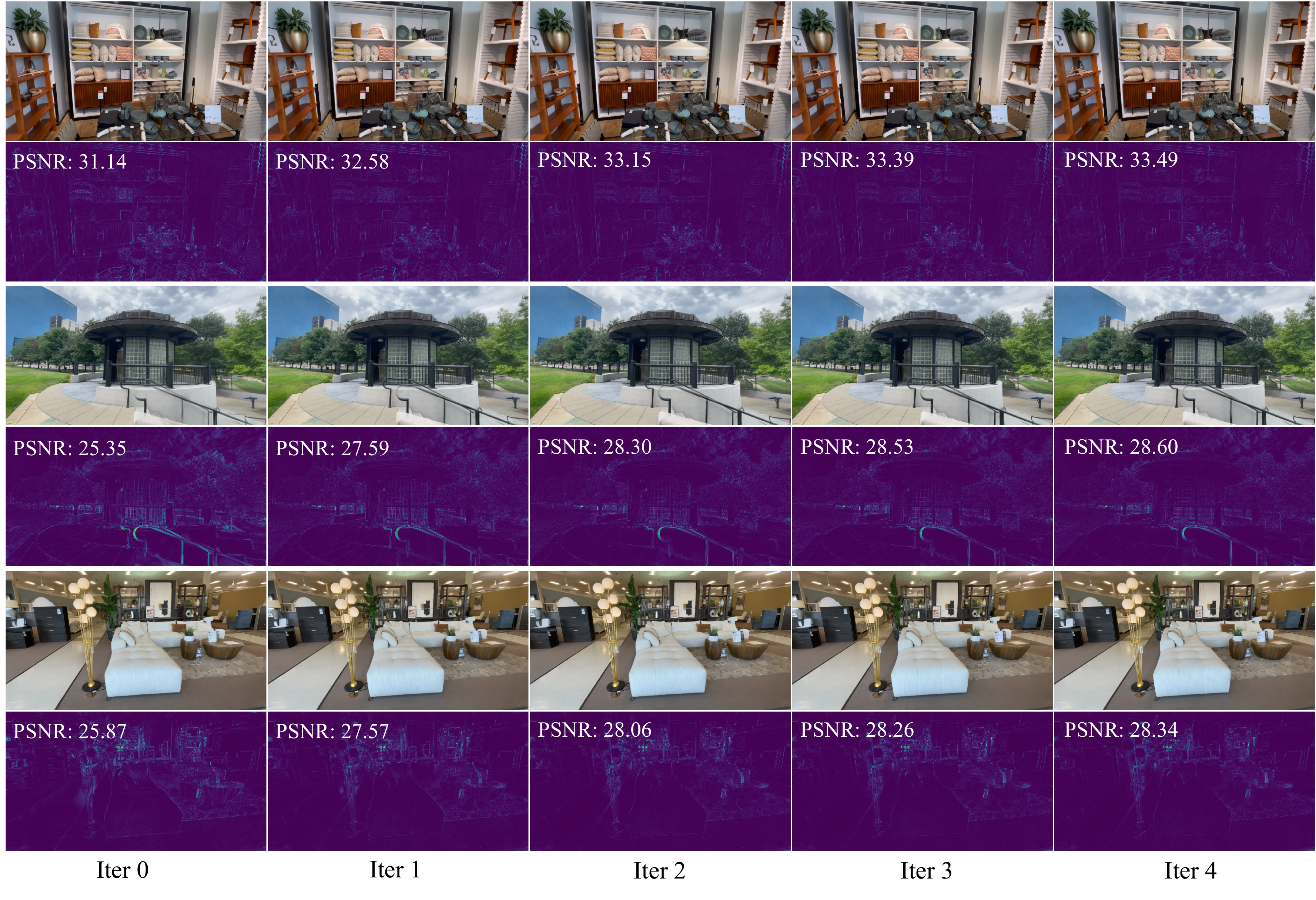}
    \vspace{-4mm}
    \caption{\textbf{Results with different numbers of iterations}. 
    }
    \label{fig:render_vs_iter}
    \vspace{-4mm}
\end{figure}

In \cref{fig:render_vs_iter}, we show the visual results with different numbers of iterations.

\begin{figure}[!t]
    \centering
    \includegraphics[width=\linewidth]{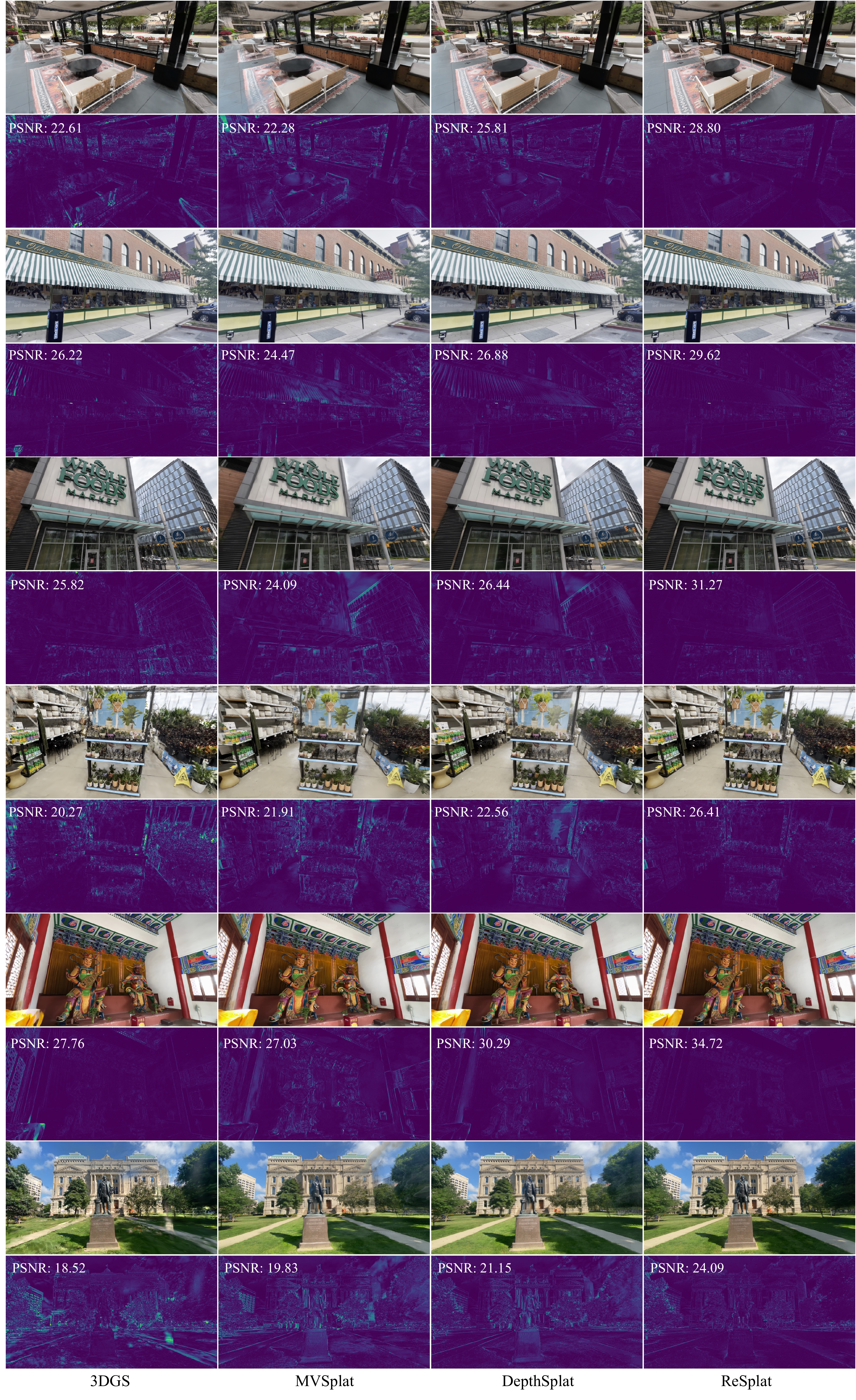}
    \vspace{-7mm}
    \caption{\textbf{More comparisons of novel view synthesis on DL3DV}.
    }
    \label{fig:vis_compare_supp}
\end{figure}

In \cref{fig:vis_compare_supp}, we show more visual comparisons with 3DGS~\cite{Kerbl2023TOG}, MVSplat~\cite{chen2024mvsplat} and DepthSplat~\cite{xu2025depthsplat} on the DL3DV dataset.

\begin{figure}[!h]
    \centering
    \includegraphics[width=0.98\linewidth]{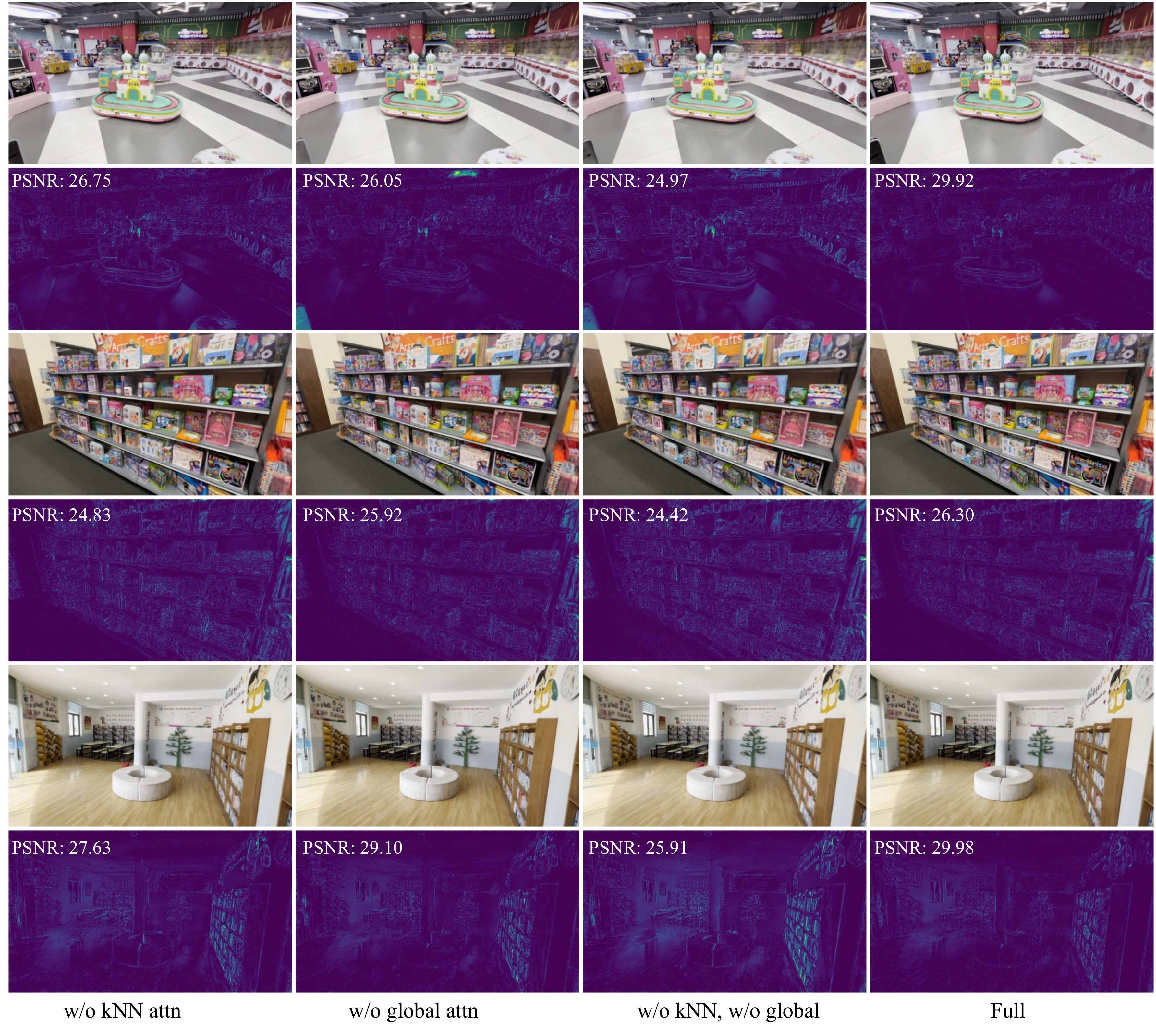}
    \vspace{-4mm}
    \caption{\textbf{Ablation of the initial model}.
    }
    \label{fig:abaltion_init}
    \vspace{-4mm}
\end{figure}

In \cref{fig:abaltion_init}, we show the visual ablation results of our initial model.

\begin{figure}[!h]
    \centering
    \includegraphics[width=0.98\linewidth]{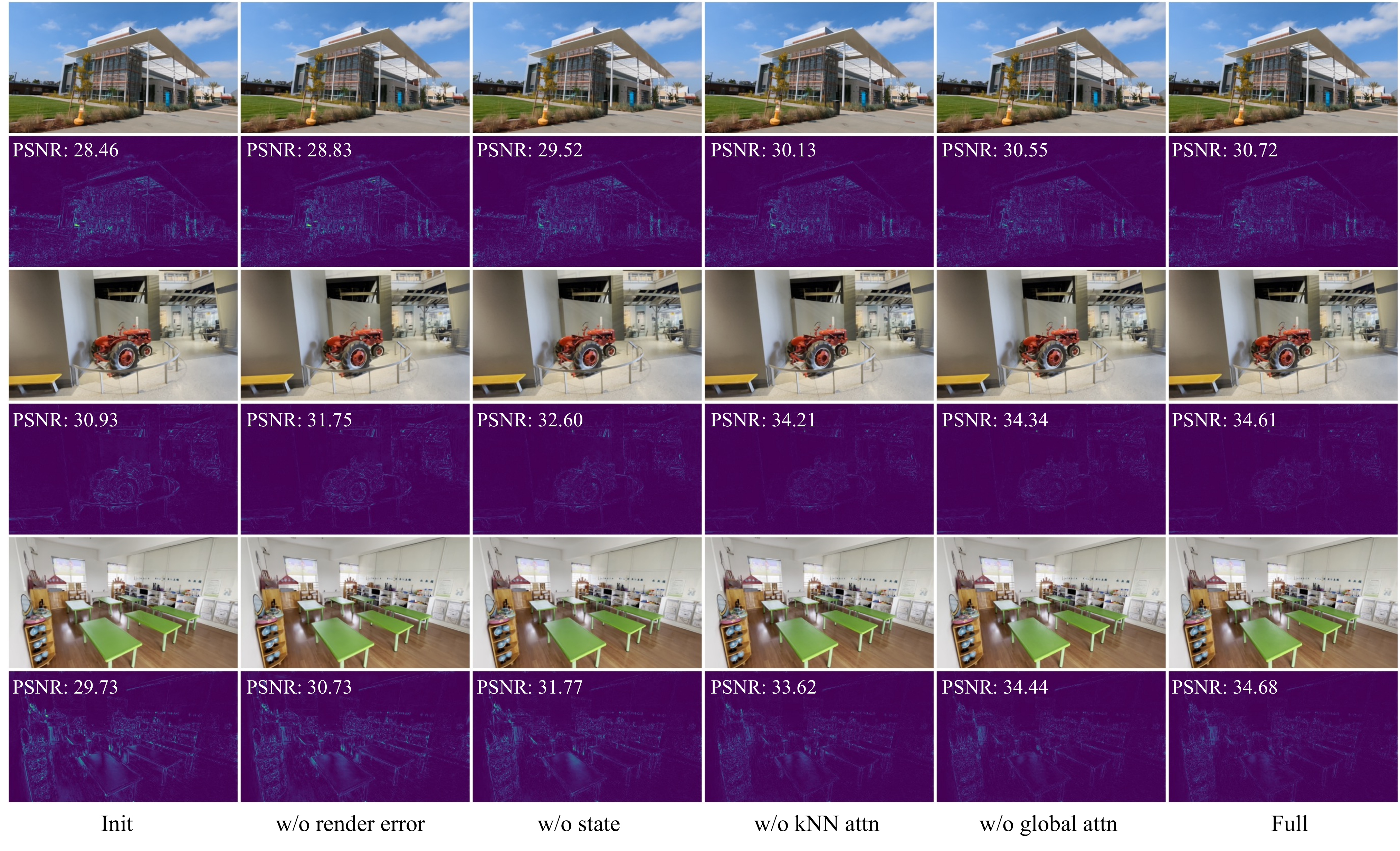}
    \vspace{-4mm}
    \caption{\textbf{Ablation of the recurrent model}.
    }
    \label{fig:abaltion_recurrent}
    \vspace{-4mm}
\end{figure}

In \cref{fig:abaltion_recurrent}, we show the visual ablation results of our recurrent model.

\end{document}